\documentclass[letterpaper, 10 pt, conference]{ieeeconf}  % Comment this line out if you need a4paper

\IEEEoverridecommandlockouts                              % This command is only needed if 
\usepackage{balance}
\usepackage{color}

\usepackage{graphics} % for pdf, bitmapped graphics files
\usepackage{amsmath,amssymb,amsfonts} % assumes amsmath package installed
\usepackage{bm}
\usepackage{subcaption}	 
\usepackage{graphicx}
\usepackage{multirow}
\usepackage{array}

\usepackage{url}
\makeatletter
\let\NAT@parse\undefined
\makeatother
\usepackage{hyperref}
\usepackage[capitalize,nameinlink]{cleveref}
\usepackage{xcolor}

\usepackage{float}
\usepackage{booktabs}
\usepackage[ruled,vlined]{algorithm2e}
\usepackage{multirow}
\usepackage{textcomp}
\usepackage{setspace}
\usepackage{tabularx}
\usepackage{makecell}
\usepackage{colortbl}  
\usepackage{pifont}

\definecolor{ceiling}{RGB}{214, 122, 122}
\definecolor{floor}{RGB}{122, 214, 122}
\definecolor{wall}{RGB}{122, 122, 214}
\definecolor{window}{RGB}{214, 214, 122}
\definecolor{chair}{RGB}{214, 122, 214}
\definecolor{bed}{RGB}{161, 222, 245}
\definecolor{sofa}{RGB}{201, 173, 217}
\definecolor{table}{RGB}{245, 184, 122}
\definecolor{tvs}{RGB}{153, 184, 122}
\definecolor{furniture}{RGB}{122, 184, 184}
\definecolor{objects}{RGB}{82, 138, 199}

\graphicspath{{ieeeconf_2/Raster_Vla/}}

\makeatletter
\@ifundefined{IEEEkeywords}{\newenvironment{IEEEkeywords}{\par\noindent\ignorespaces}{\par}}{}
\makeatother

\makeatletter
\let\raloldcaption\caption
\newcommand{\ralfigcaptionspace}{\ifx\@captype\figure\vspace{5pt}\fi}
\renewcommand{\caption}{\@ifnextchar[{\ralcaptionwith}{\ralcaptionwithout}}
\newcommand{\ralcaptionwithout}[1]{\ralfigcaptionspace\raloldcaption{#1}}
\newcommand{\ralcaptionwith}[2][]{\ralfigcaptionspace\raloldcaption[#1]{#2}}
\makeatother

\makeatletter
\def\input@path{{ieeeconf_2/Raster_Vla/}{ieeeconf_2/Raster_Vla/sections/}{ieeeconf_2/Raster_Vla/figures/}{ieeeconf_2/Raster_Vla/tables/}{ieeeconf_2/Raster_Vla/supplementary/}{ieeeconf_2/Raster_Vla/supplementary/sections/}{ieeeconf_2/Raster_Vla/supplementary/figures/}{ieeeconf_2/Raster_Vla/supplementary/tables/}}
\makeatother

\title{\LARGE \bf
DriveStack-VLA: Render-Teacher Alignment for BEV-Based DeepStack Vision-Language-Action Model
}

\author{
Jingke Wang$^{1,*}$, Zhenru Zhao$^{1,*}$, Shuangming Lei$^{1}$, Hao Su$^{1}$, Yuehao Huang$^{1}$, Yijia Xie$^{1}$, Kai Tang$^{1}$, \\
Guanglin Xu$^{2}$, AiXue Ye$^{2}$, Yukai Ma$^{1,\dagger}$, Yong Liu$^{1,\dagger}$
\thanks{$^{1}$ Zhejiang University, Hangzhou, China.}%
\thanks{$^{2}$ The 2012 Labs, Huawei.}%
\thanks{$^*$ These authors contributed equally to this work.}%
\thanks{$^{\dagger}$ The corresponding author.}%
\thanks{This work has been submitted to the IEEE for possible publication. Copyright may be transferred without notice, after which this version may no longer be accessible.}
}%
\begin{document}

\maketitle
\thispagestyle{empty}
\pagestyle{empty}

%%%%%%%%%%%%%%%%%%%%%%%%%%%%%%%%%%%%%%
% Main 
% \begin{abstract}
% Vision-Language-Action driving models convert a pretrained Vision-Language Model into a driving policy, allowing them to use world knowledge and follow language guidances.
% However, existing VLA driving models still suffer from weak visual geometry modeling, insufficient scenario diversity and inadequate coverage of key perceptual cues in expert demonstration data, and limited practical support for multimodal trajectory outputs.
% In this paper, we present \textbf{DriveStack-VLA}, a framework built upon a large VLM backbone. 
% To address the geometric and perceptual limitations, we inject a Bird-Eye-View representation into the Large Language Model decoder through a DeepStack-style connection, and propose Render-Teacher Alignment to align the perceptual focus of real images with that of rasterized images. Furthermore, we align the policy using Group Relative Policy Optimization with a joint reward consisting of a driving metric and a strict format constraint. Finally, to bridge the gap in multimodal trajectory selection, we introduce a head-based self-critique module that ranks sampled trajectories and conditionally refines the best one.
% DriveStack-VLA achieves 91.6 PDMS on NAVSIMv1, 91.0 EPDMS on NAVSIMv2 (with the human penalty filter enabled), and a driving score of 79.49 with a success rate of 56.36\% on the closed-loop Bench2Drive.
% More visualizations are available on our project page: \url{https://anonymous.4open.science/w/drivestack-vla/}.
% \end{abstract}

\begin{abstract}
Vision-Language-Action driving models convert a pretrained Vision-Language Model into a driving policy, allowing them to use world knowledge and follow language guidances.
However, existing VLA driving models still lack driving-oriented spatial intelligence: their policies are mainly grounded on perspective image tokens and language priors, while precise motion planning requires metric geometry, top-down scene structure, and attention to safety-critical perceptual cues.
This limitation makes current models vulnerable to weak visual geometry modeling and perceptual coverage in expert demonstrations.
In this paper, we present \textbf{DriveStack-VLA}, a framework built upon a large VLM backbone.
To strengthen the spatial grounding of VLA driving, we develop dual visual modeling components. 
We inject a Bird-Eye-View representation into the Large Language Model decoder through a DeepStack-style connection, and propose Render-Teacher Alignment to align the perceptual focus of real images with that of rasterized images. Furthermore, to bridge the gap in multimodal trajectory selection, we introduce a head-based self-critique module that ranks sampled trajectories and conditionally refines the best one.
DriveStack-VLA achieves 91.6 PDMS on NAVSIMv1, 91.0 EPDMS on NAVSIMv2 (with the human penalty filter enabled), and a driving score of 79.49 with a success rate of 56.36\% on the closed-loop Bench2Drive.
More visualizations are available on our project page: \url{https://anonymous.4open.science/w/drivestack-vla/}.
\end{abstract}


\section{Introduction}
\label{sec:intro}

End-to-end Vision-Language-Action (VLA) driving models are designed to employ multimodal large language models to ingest multi-view images and linguistic commands for direct future trajectory prediction~\cite{fu2025orion,zeng2025futuresightdrive}. The prevalence of such model paradigms stems from the strong world modeling and reasoning proficiency gained through large-scale pre-training in modern Large Language Models (LLMs)~\cite{brown2020language,guo2025deepseek} and Vision-Language Models (VLMs)~\cite{li2023blip,bai2025qwen3}.
% Recent advances in Large Language Models (LLMs)~\cite{brown2020language,guo2025deepseek} and Vision-Language Models (VLMs)~\cite{li2023blip,bai2025qwen3} have demonstrated robust world knowledge and reasoning capabilities derived from large-scale pretraining.
% This progress motivates the development of Vision-Language-Action (VLA) end-to-end driving models~\cite{fu2025orion,zeng2025futuresightdrive}, where a multimodal LLM processes multi-view images and language goals to generate future trajectories directly.

\begin{figure*}[t]
      \centering
       % 1. 压减图片与 Caption 之间的距离
      \setlength{\abovecaptionskip}{4pt} 
      % 2. 压减 Caption 与下方正文之间的距离
      \setlength{\belowcaptionskip}{-12pt}
      \includegraphics[width=\linewidth]{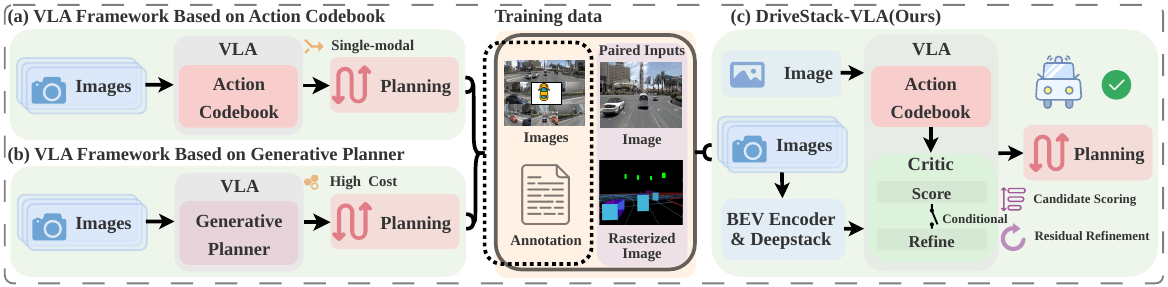}
      % \vspace{-1pt}
      \captionsetup{font={small}}
      \caption{\textbf{The difference between DriveStack-VLA and other paradigms.} Compared to other paradigms, our VLA-based method improves both the data and model sides: it enhances visual supervision to mitigate inadequate coverage of key perceptual cues, and covariate shift during SFT, injects a DeepStack-style BEV feature to strengthen geometric grounding, and equips a critic that selects and conditionally refines the best trajectory from candidates.
      }
      \label{fig:intro}
      \vspace{-5pt} 
\end{figure*}

However, a reliable driving policy cannot be built on language-level reasoning alone.
Autonomous driving is inherently a spatial decision-making problem: the policy must localize agents and lanes, reason about metric distances and future motions, and select a trajectory under strict safety constraints.
This requires driving-oriented spatial intelligence, where semantic cues from images are organized into a geometry-aware scene representation that can directly support action decoding.
Existing VLA driving models still fall short of this requirement.
Most methods~\cite{fu2025orion,li2025recogdrive,zhou2025autovla,xu2025wam,li2026sgdrive} represent scenes primarily with 2D perspective image tokens.
Although such tokens preserve rich appearance information, they are viewpoint-dependent, redundant across cameras and histories, and only implicitly encode the top-down geometry needed by planning.
% As a result, action tokens may rely on high-level visual-language correlations without being sufficiently grounded in metric road layout, agent positions, and long-range spatial context.
% Incorporating more views or longer histories further increases token sequence length, causing substantial computational overhead.
% However, turning a VLM into a reliable driving policy still faces three major limitations.
% First, existing works lack a tailored scene representation paradigm for driving-oriented VLA systems. Most VLA driving models~\cite{fu2025orion,li2025recogdrive,zhou2025autovla,xu2025wam,li2026sgdrive} represent scenes primarily with 2D image tokens, which results in fragile geometric grounding; furthermore, incorporating additional views or extended histories leads to a rapid increase in token sequence length and thus incurs substantial computational overhead. 
% Although Bird-Eye-View (BEV) representations~\cite{li2024bevformer} alleviate this limitation in conventional end-to-end driving pipelines, they fail to adapt to VLA frameworks due to the absence of a dedicated projection module that maps BEV features into LLM latent space.
This missing spatial grounding also affects how VLA policies learn from data.
Driving-language datasets and custom instruction paradigms have been shown effective for enhancing model capability~\cite{sima2024drivelm}, but expert demonstrations alone provide limited coverage of rare, safety-critical, and recovery-oriented scenarios.
Rasterized or rendered views offer a scalable way to expose models to cleaner geometric structures and more diverse counterfactual situations~\cite{feng2025rap}; nevertheless, directly mixing them with real images introduces a new challenge.
For VLA driving, the key issue is not only whether synthetic visual features look similar to real images, but whether the action decoder attends to the same planning-relevant cues under both domains.
Without such alignment, the model may benefit from augmented samples during training while still relying on fragile or distracting visual evidence at deployment.

To address these limitations, we adopt DeepStack-style connections~\cite{meng2024deepstack} to inject internally constructed BEV representations into the LLM decoder, delivering robust geometric priors for long-range driving contexts. Based on this paradigm, we propose \textbf{DriveStack-VLA}, a three-stage training framework integrating a multimodal LLM planner and an enhanced visual stack.

% 2) We also use 3d rasterized renderings as a visual augmentation signal because they are controllable and emphasize safety-critical structure. RAP shows that rasterized renderings can be an efficient alternative to heavier 3D generation pipelines~\cite{feng2025rap}, which are often built on costly rendering or novel-view synthesis tools such as NeRF or Gaussian splatting~\cite{mildenhall2021nerf,kerbl20233d}. At the same time, naive feature alignment to raster renderings in a multimodal LLM can be dominated by the black background, and adversarial domain alignment can be unstable in large-scale training~\cite{ganin2016domain}. Moreover, the key issue is not only the appearance of visual features, but also how planning tokens attend to visual tokens. We therefore treat renderings as a teacher modality and transfer supervision to real images through masked camera-token alignment and action-to-vision attention distillation~\cite{zagoruyko2016paying}, which preserves the benefit of raster augmentation without introducing an additional adversarial domain head~\cite{ganin2016domain}. 
\textbf{Stage 1: Supervised Fine-Tuning (SFT)}
, we employ controllable rasterized images to augment the image dataset for the training. Prior work~\cite{feng2025rap} indicates that 3D rasterization can provide an efficient alternative to heavier 3D generation pipelines~\cite{mildenhall2021nerf,kerbl20233d} that rely on costly rendering or novel-view synthesis. For VLA driving models, the key issue is not only the appearance of visual features, but also how action tokens attend to visual tokens. We therefore align the perceptual focus of the real image with that of the rasterized image, which preserves the benefit of raster augmentation without introducing an additional adversarial domain head~\cite{ganin2016domain}. 
\textbf{Stage 2: Reinforcement Fine-Tuning (RFT)}, we align the proposal distribution for sampling-based driving through a Group Relative Policy Optimization~\cite{shao2024deepseekmath} (GRPO) objective. This optimization integrates a driving reward with a format reward, thereby guaranteeing the generation of strictly decodable action tokens.
\textbf{Stage 3: Scoring and Refinement}
Built on LLM decoder hidden states, lightweight scoring and refinement heads rank candidate trajectories and optimize selected outputs. This pipeline avoids redundant text generation and eliminates slow iterative generate-critique-rewrite loops during inference.
% Finally, we train lightweight scoring and refinement heads \textbf{in Stage-3 (scoring and refinement)} based on the hidden states of the LLM decoder to rank the sampled trajectories and to refine the selected trajectory when necessary without extra text generation at inference time.

% We extensively evaluate DriveStack-VLA on real-world and simulated benchmarks, including NAVSIMv1/v2~\cite{dauner2024navsim,cao2025pseudo} and Bench2Drive~\cite{jia2024bench2drive}. The results demonstrate that DriveStack-VLA achieves superior performance across end-to-end autonomous driving benchmarks and validate that the proposed three-stage training recipe significantly enhances planning capabilities. The main contributions of this paper are summarized as follows:
We extensively evaluate DriveStack-VLA on real-log-driven NAVSIMv1/v2~\cite{dauner2024navsim,cao2025pseudo} and simulated closed-loop Bench2Drive~\cite{jia2024bench2drive}. The results demonstrate that DriveStack-VLA achieves superior performance across end-to-end autonomous driving benchmarks. The main contributions of this paper are summarized as follows:
\begin{itemize}
\item We propose \textbf {DriveStack-VLA}, a unified three-stage VLA driving framework addressing geometric, dataset and trajectory sampling bottlenecks.
\item We develop dual visual modeling components: BEV DeepStack injection embeds top-down geometric priors into the LLM decoder, and Render-Teacher Alignment aligns visual-action attention across real and rasterized images.
\item We design a lightweight self-critic module for trajectory ranking and refinement. Comprehensive experiments on NAVSIMv1/v2 and Bench2Drive further validate that our method achieves state-of-the-art quantitative results on both real-log open-loop and simulated closed-loop driving benchmarks.

% \item We introduce BEV DeepStack injection for VLA driving model, which injects a top-down geometry prior into the LLM decoder under multi-view images.

% \item We propose Render-Teacher Alignment for VLA driving model, utilizing masked camera-token alignment and action-to-vision attention distillation to align the perceptual focus of the real image with that of the rasterized image.

% \item To bridge the gap in the sampling and selection of multimodal trajectories under the paradigm of action tokenization, we introduce a head-based self-critic module that effectively ranks the sampled candidates and conditionally refines the optimal trajectory.
% \item To bridge the gap in sampling and selecting multimodal trajectories under action tokenization, we introduce a head-based self-critic module that ranks sampled candidates and conditionally refines the optimal trajectory.
\end{itemize}
% Stage-1 coverage of safety-critical cues through render-based visual augmentation, while Render-Teacher Alignment transfers render supervision to real images without being dominated by background bias; 

% Stage-2 further calibrates the actor with GRPO-style reward-aligned RFT using verifiable planning rewards and strict format constraints, reducing the mismatch between logged training data and evaluation rollouts; 
% Stage-2 further calibrates the actor for best-of-$N$ sampling via a GRPO-style reward-aligned fine-tuning step, directly optimizing a verifiable EPDMS planning reward together with a strict format-consistency constraint to ensure decodable VQ-token trajectories under stochastic decoding;

% Stage-2 further calibrates the VLA for best-of-$N$ sampling via a GRPO-style reward-aligned fine-tuning step, directly optimizing a verifiable EPDMS planning reward together with a strict format-consistency constraint to ensure decodable action-token trajectories under stochastic decoding;

% and Stage-3 adds a head-based self-critique module that ranks multiple sampled trajectories and applies residual refinement when needed, leading to more reliable and accurate planning decisions.

\section{Related Works}

\subsection{End-to-End Autonomous Driving}
End-to-end driving models learn a direct mapping from raw sensor inputs to future trajectories or controls~\cite{chitta2022transfuser}. Unified architectures such as UniAD~\cite{hu2023planning} and VAD~\cite{jiang2023vad} train perception, prediction, and planning in one model and improve open-loop planning quality. SparseDrive~\cite{sun2025sparsedrive} further optimizes path generation through a multimodal trajectory-prediction framework. DiffusionDrive~\cite{liao2025diffusiondrive} and Goalflow~\cite{xing2025goalflow} use generative trajectory models to cover multiple valid futures. These methods reach strong open-loop planning scores, but offline imitation still suffers from covariate shift in closed-loop driving, and offers limited support for language goals and world knowledge.

\subsection{Vision-Language-Model in Autonomous Driving}

Although end-to-end driving models learn effectively from logged trajectories, integrating world knowledge and language goals remains challenging. This limitation motivates the application of VLMs in autonomous driving, shifting the field from language-only interpretation to unified vision-language-action policies. 
% Early methods~\cite{xu2024drivegpt4,sima2024drivelm} primarily utilize VLMs for scene description and visual question answering. While this approach improves explainability, it does not produce executable actions. Subsequent approaches decouple high-level decision generation from downstream planning. This modular design facilitates integration but prevents end-to-end optimization~\cite{liu2025vlm}. 
Recent end-to-end VLA models map multi-camera images and language goals directly to future motion. Within this paradigm, existing methods typically follow three directions: decoding continuous trajectories directly~\cite{zeng2025futuresightdrive}, incorporating generative planners to capture multimodal futures~\cite{fu2025orion,li2025recogdrive,li2026sgdrive,xu2025wam}, or discretizing motion into action tokens~\cite{zhou2025autovla}. For model training, RFT is increasingly adopted alongside SFT to align outputs with planning metrics. For instance, AutoVLA~\cite{zhou2025autovla} applies GRPO-style fine-tuning to an autoregressive policy, while WAM-Diff~\cite{xu2025wam} utilizes online GSPO. Conversely, ReCogDrive~\cite{li2025recogdrive} and SGDrive~\cite{li2026sgdrive} apply reinforcement learning exclusively to a separate diffusion planner. This configuration prevents the reward signal from directly optimizing the VLM and maintains reliance on auxiliary modules. In this work, we adopt the paradigm of discretizing planning into action tokens. Furthermore, we apply GRPO-style fine-tuning directly to the actor, utilizing a verifiable planning metric and a format-consistency reward for alignment.

% \subsection{Rendering for autonomous driving.}
% Synthetic view generation is a common approach for augmenting recovery data beyond the logged trajectory. Neural rendering methods, such as NeRF \cite{mildenhall2021nerf} and Gaussian Splatting \cite{kerbl20233d}, can replay scenes with high visual fidelity, but these methods remain slow and difficult to scale. 
% RAP \cite{feng2025rap} demonstrates that lightweight 3D rasterization can generate controllable views that emphasize lanes, agents, and signals, and that it aligns raster features with real-image features in a shared feature space.
% % For VLA driving models, the primary supervision signal includes not only image features, but also the attention patterns between action tokens and visual tokens in the LLM decoder. 
% In this work, we treat rasterized images as a teacher modality and transfer supervision to real images via masked camera-token alignment and action-to-vision attention distillation \cite{zagoruyko2016paying}. 
% This design provides a direct and efficient way to couple raster augmentation with planning, encouraging the LLM decoder
% to attend to safety-critical cues during planning.
% to align the perceptual focus of the real image with that of the rasterized image
\begin{figure*}[t]
      \centering
       % 1. 压减图片与 Caption 之间的距离
      \setlength{\abovecaptionskip}{4pt} 
      % 2. 压减 Caption 与下方正文之间的距离
      \setlength{\belowcaptionskip}{-16pt}
      \includegraphics[width=\linewidth]{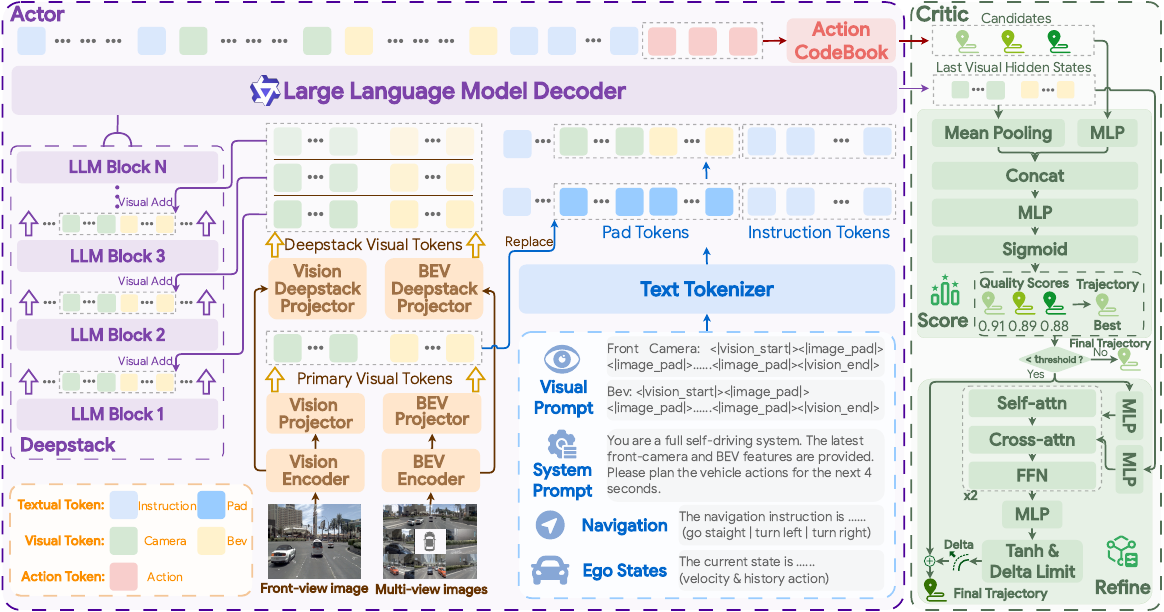}
      % \vspace{-1pt}
      \captionsetup{font={small}}
      \caption{\textbf{Architecture of DriveStack-VLA.} Built upon a VLM backbone, our actor-critic framework processes multi-view images, instructions, and ego states. The actor injects BEV features into the LLM decoder through a DeepStack-style connection to generate action-token sequences, which a frozen codebook decodes into continuous trajectories. The critic comprises two heads: a scoring head that reuses last-layer visual hidden states to rank candidates, and a refinement head that conditionally predicts bounded residuals via attention blocks if the highest score is below a threshold.
      }
      \label{fig:framework}
      \vspace{-5pt} 
\end{figure*}

\section{Methodology}
\label{sec:method}

\subsection{Framework Overview}
% \paragraph{Drivestack-VLA框架组成.}
% 我们提出的DriveStack-VLA框架由两部分组成,as show in Fig.~\ref{fig:intro}, 1)Actor(VLA Driving Model): 该模块接收多视角相机图像$\mathbf{I}=\{I_c\}_{c=1}^{C}$,其中C为相机数量，和navigation指令$\mathbf{x}以及自车状态$\mathbf{u}$作为输入, We write the full observation as $\mathbf{o}=(\mathbf{x},\mathbf{I},\mathbf{u})$.  We use Qwen3-VL~\cite{bai2025qwen3} as the vlm backbone, actor包含视觉与BEV双分支编码器，并通DeepStack injection 将视觉与bev特征逐层注入到LLM解码器中, LLM解码器以自回归的方式输出动作tokens，并通过Action Codebook解码为符合物理可行性的候选轨迹。 2)Critic (Head-based Self-Critique Module): 该模块由轻量级的打分器和细化器构成, 它复用Actor的LLM解码器的最后一层的视觉隐藏状态，并结合候选轨迹作为输入。首先，打分器对候选轨迹进行评估并输出标量Quality Scores以进行排序, 当最高得分的轨迹低于预设阈值时，条件触发细化器，利用注意力机制（Self-attn & Cross-attn）结合视觉上下文预测有界的轨迹残差$\Delta$，最终输出经过轻量级修正的安全规划轨迹。
\subsubsection{Components of DriveStack-VLA}
DriveStack-VLA contains an actor and a critic, as shown in \cref{fig:framework}.
The actor takes multi-view images $\mathbf{I}=\{I_c\}_{c=1}^{C}$, a navigation instruction $\mathbf{x}$, and ego states $\mathbf{u}$ as inputs, where $C$ is the number of cameras.
We write the full observation as $\mathbf{o}=(\mathbf{x},\mathbf{I},\mathbf{u})$.
The VLM backbone is Qwen3-VL-4B~\cite{bai2025qwen3}.
The actor includes a perspective-view vision encoder and an additional BEV encoder.
The BEV branch produces a top-down representation and injects it into the LLM decoder through a DeepStack-style connection~\cite{meng2024deepstack}.
The decoder then generates action-token sequences, and a frozen action codebook decodes these sequences into physically valid continuous trajectories.

The critic reuses the last-layer hidden states of the LLM decoder and combines them with candidate trajectories.
A lightweight scoring head outputs a scalar quality score for each candidate trajectory and ranks the candidates.
If the best score is below a preset threshold, a refinement head predicts a bounded trajectory residual with attention over visual context.

% \paragraph{训练与测试流程.}
% DriveStack-VLA 训练分成三阶段进行，As illustrated in Fig.~\ref{fig:intro}.
% Stage-1 trains the actor by supervised fine-tuning with BEV DeepStack injection and render-teacher alignment (Sec.~\ref{sec:bev}--\ref{sec:rgm3}) 冻结预训练好的vision encoder和bev encoder，训练主干llm和vision和bev投影层.
% Stage-2 冻结投影层，训练主干llm， applies reinforcement fine-tuning to improve best-of-$N$ sampling (Sec.~\ref{sec:rft}).
% Stage-3 freezes the actor and trains lightweight heads for scoring and residual refinement (Sec.~\ref{sec:heads}).

% At test time, we first greedily decode one action tail.
% Then we sample extra candidates by sampling action tokens while keeping the prefix prompt fixed.
% We rank candidates with the score head. If all scores are below a threshold, we refine the candidate closest to the threshold; otherwise we directly take the best-scoring candidate.

\subsubsection{Training Pipeline}
\begin{figure*}[t]
      \centering
       % 1. 压减图片与 Caption 之间的距离
      \setlength{\abovecaptionskip}{4pt} 
      % 2. 压减 Caption 与下方正文之间的距离
      \setlength{\belowcaptionskip}{-8pt}
      \includegraphics[width=\linewidth]{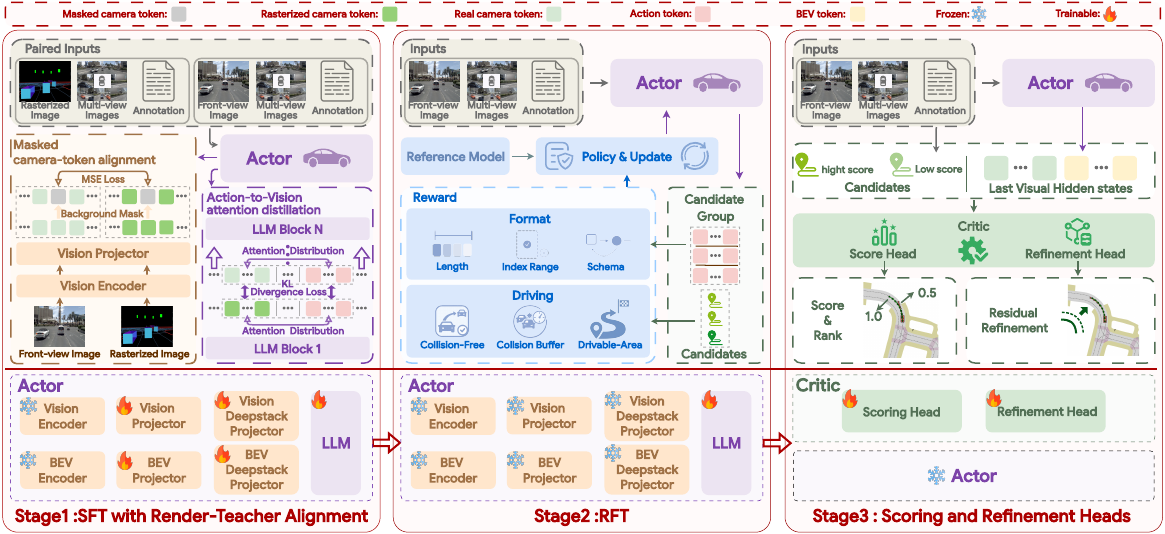}
      % \vspace{-1pt}
      \captionsetup{font={small}}
      \caption{
      \textbf{Training pipeline of DriveStack-VLA.} Stage 1 executes SFT via Render-Teacher Alignment, incorporating masked camera-token alignment and action-to-vision attention distillation. Stage 2 applies RFT utilizing a GRPO objective to align the distribution of proposals. Stage 3 freezes the actor to train lightweight scoring and refinement heads, thereby enabling candidate ranking and residual refinement.
      }
      \label{fig:training}
      \vspace{-10pt} 
\end{figure*}

Training proceeds in three stages, as shown in \cref{fig:training}.
In Stage-1, we perform SFT to train the actor with BEV DeepStack injection (\cref{sec:bev}) and Render-Teacher Alignment (\cref{sec:rgm3}).
In this stage, we freeze the pretrained vision encoder and the BEV encoder, and we train the language-model decoder together with the visual projection layers that connect camera and BEV features to the decoder.
In Stage-2, we freeze the projection layers and apply RFT to the actor with a GRPO objective to improve best-of-$K$ sampling (\cref{sec:rft}).  
In Stage-3, we freeze the actor and train lightweight scoring and refinement heads (\cref{sec:heads}).

% At inference time, the actor first decodes one candidate action tail with greedy decoding.
% It then samples additional candidates by sampling action tokens while keeping the same prefix prompt fixed.
% The score head ranks all candidates.
% If the highest score is below a threshold $\delta_s$, the refinement head refines the candidate with a score closest to $\delta_s$; otherwise, the system directly selects the highest-scoring candidate.

% At inference time, the actor samples a set of candidate action tails by sampling action tokens. The scoring head ranks all candidates. If the highest score is below a threshold $\delta_s$, the refinement head refines the candidate with a score closest to $\delta_s$; otherwise, the system directly selects the highest-scoring candidate.

% \paragraph{Action Tokenization.}
% We keep the output$\mathbf{y}=\mathbf{a}$, where $\mathbf{a}$ is a fixed-length action tail.
% We discretize a trajectory into $S$ VQ segments（这里需要找一个vqvae作为引用）. 这里S为4，代表未来4秒的轨迹范围覆盖， For each segment $s$, the actor predicts a scale token $\langle \text{scale}_{q_s}\rangle$ and a code token $\langle \text{traj}_{c_s}\rangle$.
% So the action tail is
% \begin{equation}
% \mathbf{a}=[\langle \text{scale}_{q_1}\rangle,\langle \text{traj}_{c_1}\rangle,\dots,
%            \langle \text{scale}_{q_S}\rangle,\langle \text{traj}_{c_S}\rangle].
% \label{eq:action_tail}
% \end{equation}
% A frozen VQ decoder maps (即Action Codebook) $\mathbf{a}$ to a continuous trajectory
% $\boldsymbol{\tau}=\{(x_t,y_t,\psi_t)\}_{t=1}^{T}$其中,xt、yt为相对第0s的位移，psi_t为朝向角，T为时间步.
\subsubsection{Action Tokenization.}
The output of the LLM decoder is a fixed-length action tail $\mathbf{a}$.
A continuous trajectory is discretized into $S$ vector-quantized segments using a codebook in the style of vector-quantized autoencoders~\cite{van2017neural}.
% Here, $S=4$, which corresponds to a $4$ seconds planning horizon in our setup.
Here, $S$ denotes the planning horizon in seconds.
For each segment $s\in\{1,\dots,S\}$, the LLM decoder predicts one scale token $\langle \text{scale}_{q_s}\rangle$ and one code token $\langle \text{traj}_{c_s}\rangle$, 
% $\mathbf{a}=
% [\langle \text{scale}_{q_1}\rangle,\langle \text{traj}_{c_1}\rangle,\dots,\langle \text{scale}_{q_S}\rangle,\langle \text{traj}_{c_S}\rangle]$, 
% \begin{equation}
% \mathbf{a}=
% [\langle \text{scale}_{q_1}\rangle,\langle \text{traj}_{c_1}\rangle,\dots,\langle \text{scale}_{q_S}\rangle,\langle \text{traj}_{c_S}\rangle],
% \label{eq:action_tail}
% \end{equation}
where $q_s$ is the discrete index of a scale token for segment $s$, and $c_s$ is the discrete index of a trajectory code token for segment $s$.
A frozen vector-quantized decoder, which we refer to as the action codebook, maps $\mathbf{a}$ to a continuous trajectory
$\boldsymbol{\tau}=\{(x_t,y_t,\psi_t)\}_{t=1}^{T}$.
Here, $T$ is the number of trajectory time steps, $(x_t,y_t)$ is the planar displacement at time step $t$ relative to the ego pose at $t=0$, and $\psi_t$ is the yaw angle at time step $t$.

\subsection{BEV DeepStack Injection}
\label{sec:bev}
%这里缺少简要解释、动机，为什么要进行bev deepstack injection
% Qwen3-VL natively supports multi-image token grids. We add a BEV encoder $E_{\text{bev}}$ (BEVFormer~\cite{li2024bevformer})
% to get a top-down feature map from driving context, and project it to BEV tokens:
% \begin{equation}
% \mathbf{Z}^{\text{bev}}=P_{\text{bev}}\!\left(E_{\text{bev}}(\mathbf{I},\mathbf{p})\right)\in\mathbb{R}^{N_b\times d},
% \label{eq:bev_tokens}
% \end{equation}
% where $\mathbf{p}$ 是相机参数, $N_b$ is the BEV token count and $d$ is the decoder 维度.
% Perspective-view image tokens provide rich appearance cues, but geometric grounding can be weak when the context is long or when many views are present.
% A BEV representation helps because it provides a top-down geometry prior.
% A simple approach is to concatenate BEV tokens to the input token sequence, but this can be unstable under long sequences and can increase the effective context length.
% We instead inject BEV information through a DeepStack-style multi-level pathway so the decoder can use BEV cues in a stable way without adding many extra tokens at the input.
Although perspective camera tokens provide appearance cues, geometric grounding often becomes unstable given long contexts and multiple camera views. Therefore, we introduce a BEV branch and inject the corresponding representations through a DeepStack-style connection to provide the LLM decoder with a stable top-down prior. Specifically, a BEV encoder $E_{\text{bev}}$, based on BEVFormer~\cite{li2024bevformer}, is employed to extract a top-down feature map from the driving context. This map is subsequently projected into BEV tokens as follows:
\begin{equation}
\mathbf{Z}^{\text{bev}} = P_{\text{bev}}\!\left(E_{\text{bev}}(\mathbf{I},\mathbf{p})\right)\in\mathbb{R}^{N_b\times d},
\label{eq:bev_tokens}
\end{equation}
where $\mathbf{p}$ denotes the camera parameters, $P_{\text{bev}}(\cdot)$ represents a learned projection that generates a token sequence and shares an identical architecture with the PatchMerger of Qwen3-VL~\cite{bai2025qwen3}, $N_b$ indicates the number of BEV tokens, and $d$ is the hidden dimension of the decoder.

The DeepStack interface of Qwen3-VL accommodates per-layer visual memory. Let $\mathcal{L}_{\text{ds}}=\{\ell_1,\dots,\ell_L\}$ denote the set of LLM decoder layers into which the DeepStack memory is injected. For each layer $\ell\in\mathcal{L}_{\text{ds}}$, a BEV pathway derives layer-specific BEV tokens $\mathbf{Z}^{\text{bev}}_{\ell}\in\mathbb{R}^{N_b\times d}$ from the shared BEV feature map. Concurrently, the vision encoder provides layer-specific camera tokens $\mathbf{Z}^{\text{cam}}_{\ell}\in\mathbb{R}^{N_c\times d}$, where $N_c$ represents the number of camera tokens per view at layer $\ell$ after pooling. The per-layer DeepStack memory is then formulated as $\mathbf{V}_{\ell}=[\mathbf{Z}^{\text{cam}}_{\ell};\mathbf{Z}^{\text{bev}}_{\ell}]$ for $\ell\in\mathcal{L}_{\text{ds}}$, with $[\cdot;\cdot]$ denoting concatenation along the token dimension. This design successfully integrates both BEV and camera cues through a unified multi-level interface.

\subsection{Render-Teacher Alignment}
\label{sec:rgm3}
%这里需要再根据前面的介绍再简要提出强调一下为什么要进行Render-teacher alignment，为什么要加入3d rasterization
% Each sample also provides rasterized renders $\tilde{\mathbf{I}}=\{\tilde{I}_c\}_{c=1}^{C}$. 这里制作3d rasterization的过程参照了RAP的流程，模型的输入包括muti-camera images为制作bev特征使用，而额外加入了前视图作为输入vision encoder通道使用，在模型输入中使用了front camera 和 front render作为paried input，当作bs为2的输入。与RAP这种传统端到端模型不同，我们这里需要对齐的是vision tokens，而Raster renderings often have large pure-black background regions, if we align tokens everywhere, background patches dominate the gradients. So we compute a render-guided token mask and only align the foreground.
% Logged images provide realistic appearance, but safety-critical structure can be sparse and hard to cover in long-tail cases.
Rasterized images can highlight lanes, agents, and traffic signals in a controllable way, and we construct rasterized images following RAP~\cite{feng2025rap}.
% A direct feature alignment between real images and rasterized images can be dominated by large black background regions in rasterized images.
% Furthermore, for VLA driving models, the key issue is not only the appearance of visual features, but also how action tokens attend to visual tokens.
We treat rasterized images as a teacher modality to align the perceptual focus of the real image with that of the rasterized image through masked camera-token alignment and action-to-vision attention distillation~\cite{zagoruyko2016paying}.

Each training sample provides real multi-view images $\mathbf{I}=\{I_c\}_{c=1}^{C}$ and paired rasterized images $\tilde{\mathbf{I}}=\{\tilde{I}_c\}_{c=1}^{C}$.
We construct a paired input for the front-view image and treat it as a batch of size $2$. 
We use superscripts $r$ and $m$ to denote tensors from the real element and the rasterized element, respectively.
The rasterized image serves as the teacher, so gradients do not propagate through teacher-side tensors by applying $\mathrm{sg}(\cdot)$ in teacher-to-student alignment terms.
% \paragraph{Render-guided mask.}

\subsubsection{Masked Camera-Token Alignment}
A direct feature alignment between real images and rasterized images can be dominated by large black background regions in rasterized images.
We therefore compute a render-guided soft mask to down-weight background tokens during alignment. 
Let the paired images be patchified and merged into $N_v$ visual tokens.
For each token index $k\in\{1,\dots,N_v\}$, let $\bar{p}_k$ be the mean pixel value of the corresponding rasterized patch after normalization to $[-1,1]$.
We map it to $[0,1]$ and define:
\begin{equation}
w_k=\mathrm{clip}\!\left(\frac{(\bar{p}_k+1)/2}{\tau},0,1\right)^{\gamma},\quad w_k\in[0,1],
\end{equation}
where $\mathrm{clip}(x,0,1)=\min(\max(x,0),1)$, $\tau>0$ controls the foreground threshold, and $\gamma>0$ controls the sharpness.
This design assigns small weights to background patches (black) and large weights to foreground patches.

Let $\mathbf{z}^{r}_k\in\mathbb{R}^{d}$ and $\mathbf{z}^{m}_k\in\mathbb{R}^{d}$ denote the camera-token embeddings at patch $k$ from the real image and the rasterized image, after the same pooling and projection to dimension $d$.
We align them with a mask-weighted mean-squared error (MSE), and the rasterized tokens are treated as the teacher with stop-gradient $\mathrm{sg}(\cdot)$:
\begin{equation}
L_{\text{mask}}
=
\frac{\sum_{k=1}^{N_v} w_k \,\lVert \mathbf{z}^{r}_k-\mathrm{sg}(\mathbf{z}^{m}_k)\rVert_2^2}
{\sum_{k=1}^{N_v} w_k + \epsilon},
\label{eq:masked_align}
\end{equation}
where $\epsilon>0$ is a small constant for numerical stability.

% \paragraph{DeepStack camera-token alignment.}
% The vision encoder also produces DeepStack camera tokens at multiple layers.
% Let $\mathbf{Z}^{r}_{\ell}\in\mathbb{R}^{N_c\times d}$ and $\mathbf{Z}^{m}_{\ell}\in\mathbb{R}^{N_c\times d}$ be the camera DeepStack tokens from the real-image and the render at layer $\ell$.
% We align them with
% \begin{equation}
% \mathcal{L}_{\text{ds-align}}
% =
% \frac{1}{|\mathcal{L}_{\text{ds-align}}|}
% \sum_{\ell\in\mathcal{L}_{\text{ds-align}}}
% \mathrm{MSE}\!\left(\mathbf{Z}^{r}_{\ell},\,\mathrm{sg}(\mathbf{Z}^{m}_{\ell})\right).
% \label{eq:deepstack_align}
% \end{equation}
% In Equation~\eqref{eq:deepstack_align}, $\mathcal{L}_{\text{ds-align}}$ is the set of layers used for token alignment, $|\mathcal{L}_{\text{ds-align}}|$ is its size, and $\mathrm{MSE}(\cdot,\cdot)$ is the mean squared error over all token positions and channels.

\subsubsection{Action-to-Vision Attention Distillation}
% For planning, it is important to shape which visual tokens the action tokens attend to.
For planning, it is crucial to explicitly shape the attention distribution from action tokens to visual tokens.
We distill attention patterns from the rasterized image (teacher) pass to the real image pass (student).
Let $\mathbf{A}^{r,\ell,h}\in\mathbb{R}^{N\times N}$ be the self-attention matrix at LLM decoder layer $\ell$ and head $h$ from the real image pass, and let $\mathbf{A}^{m,\ell,h}\in\mathbb{R}^{N\times N}$ be the corresponding matrix from the rasterized image pass, where $N$ is the total token length in the LLM decoder context.
Let $\mathcal{Q}$ be the set of query positions that correspond to supervised action tokens in the target sequence, and let $\mathcal{K}_{\text{cam}}$ be the set of key positions that correspond to camera tokens of the paired images.
We first average attention over heads and query positions to derive a non-negative vector over camera keys:
\begin{equation}
p^{r}_{\ell}(k)
=
\mathrm{Norm}\Big(
\frac{1}{|\mathcal{Q}|}\sum_{q\in\mathcal{Q}}
\frac{1}{H}\sum_{h=1}^{H}
\mathbf{A}^{r,\ell,h}_{q\rightarrow k}
\Big),\quad k\in\mathcal{K}_{\text{cam}},
\label{eq:attn_pr}
\end{equation}
where $H$ is the number of attention heads, $\mathbf{A}^{r,\ell,h}_{q\rightarrow k}$ is the attention weight from query $q$ to key $k$, and $\mathrm{Norm}(\mathbf{v})=\mathbf{v}/\sum_i v_i$ normalizes a non-negative vector to sum to 1.
We define $p^{m}_{\ell}$ analogously from $\mathbf{A}^{m,\ell,h}$.

We then apply a temperature $T_a>0$ and compute distributions:
\begin{equation}
\begin{aligned}
\tilde{p}^{r}_{\ell}
&=\mathrm{softmax}\Big(\log(p^{r}_{\ell}+\epsilon)/T_a\Big),\\[-1mm]
\tilde{p}^{m}_{\ell}
&=\mathrm{softmax}\Big(\log(p^{m}_{\ell}+\epsilon)/T_a\Big).
\end{aligned}
\label{eq:attn_temp}
\end{equation}

Finally, we distill attention by minimizing the Kullback-Leibler (KL) divergence loss:
\begin{equation}
L_{\text{attn}}
=
\frac{1}{|\mathcal{L}_{\text{attn}}|}
\sum_{\ell\in\mathcal{L}_{\text{attn}}}
\mathrm{KL}\!\left(\mathrm{sg}(\tilde{p}^{m}_{\ell})\,\|\,\tilde{p}^{r}_{\ell}\right),
\label{eq:attn_distill}
\end{equation}
where $\mathcal{L}_{\text{attn}}$ is the set of LLM decoder layers where attention is recorded. In our implementation.

% , we set $\mathcal{L}_{\text{attn}}=\{0,1,2\}$ So attention distillation is applied only to the first three LLM decoder layers.

% \paragraph{Stage-1 loss.}
% We also apply the same token-level CE loss on the render pass (weighted) so the render branch stays task-aligned.The full Stage-1 objective:
% \begin{equation}
% \mathcal{L}_{\text{stage1}}
% =
% \mathcal{L}^{r}_{\text{CE}}
% +\lambda_{\text{meta}}\mathcal{L}^{m}_{\text{CE}}
% +\lambda_{\text{mask}}\mathcal{L}_{\text{mask}}
% +\lambda_{\text{ds}}\mathcal{L}_{\text{ds}}
% +\lambda_{\text{attn}}\mathcal{L}_{\text{attn}}.
% \label{eq:stage1_total}
% \end{equation}

% \paragraph{Stage-1 supervised loss.}
% Given a target completion $\mathbf{y}^{\star}$, Stage~1 uses a standard autoregressive cross-entropy loss:
% \begin{equation}
% \mathcal{L}_{\text{CE}}(\theta;\mathbf{o},\mathbf{y}^{\star})
% = -\sum_{t=1}^{|\mathbf{y}^{\star}|}
% \log p_{\theta}\!\left(y^{\star}_t \mid \mathbf{o}, \mathbf{y}^{\star}_{<t}\right).
% \label{eq:ce}
% \end{equation}
% % In Equation~\eqref{eq:ce}
% where $\theta$ denotes trainable parameters, $p_{\theta}(\cdot)$ is the token distribution from the decoder, and $\mathbf{y}^\star_{<t}$ denotes the length-$(t-1)$ prefix of $\mathbf{y}^\star$.

\begin{table}[tb]
\caption{Comparison with state-of-the-art methods on the \textbf{NAVSIMv1}~\cite{dauner2024navsim} (navtest). $^\dagger$ denotes models fine-tuned on the NAVSIM dataset. We additionally report results with RFT to enable fair comparison with methods that adopt such training strategies.}
\vspace{-5pt}
\label{tab:navsim_v1}
\centering
\tiny
\setlength{\tabcolsep}{1.2pt}
\renewcommand{\arraystretch}{0.92}
\resizebox{\columnwidth}{!}{%
\begin{tabular}{@{}lccccc>{\columncolor{gray!15}[\tabcolsep][0pt]}c@{}}
\toprule
Method & NC $\uparrow$ & DAC $\uparrow$ & TTC $\uparrow$ & Comf. $\uparrow$ & EP $\uparrow$ & \textbf{PDMS} $\uparrow$ \\
\midrule
Human & 100.0 & 100.0 & 100.0 &  99.9 & 87.5 & 94.8 \\
\midrule
\multicolumn{7}{@{}l@{}}{\textbf{End-to-end-based methods}} \\
TransFuser~\cite{chitta2022transfuser} & 97.7 & 92.8 & 92.8 & 100.0 & 79.2 & 84.0 \\
PARA-Drive~\cite{weng2024drive} & 97.9 & 92.4 & 93.0 & 99.8 & 79.3 & 84.0 \\
DiffusionDrive~\cite{liao2025diffusiondrive} & 98.2 & 96.2 & 94.7 & 100.0 & 82.2 & 88.1 \\
WoTE~\cite{li2025end} & 98.5 & 96.8 & 94.4 & 99.9 & 81.9 & 88.3 \\
\midrule
\multicolumn{7}{@{}l@{}}{\textbf{VLMs-based methods (SFT)}} \\
AutoVLA~\cite{zhou2025autovla} & 96.9 & 92.4 & 88.1 & 99.1 & 75.8 & 80.5 \\
Qwen2.5-VL-8B$^\dagger$~\cite{bai2025qwen25vltechnicalreport} & 97.8 & 92.1 & 92.8 & 100.0 & 78.3 & 83.3 \\
Qwen3-VL-4B$^\dagger$~\cite{bai2025qwen3} & 97.7 & 92.4 & 93.9 & 100.0 & 78.5 & 84.0 \\
ReCogDrive~\cite{li2025recogdrive} & 98.3 & 95.1 & 94.3 & 100.0 & 81.1 & 86.8 \\
SGDrive~\cite{li2026sgdrive} & 98.6 & 95.1 & \textbf{95.4} & 100.0 & 81.2 & 87.4 \\
\midrule
\textbf{DriveStack-VLA (ours)} & \textbf{98.8} & \textbf{97.2} & 94.9 & \textbf{100.0} & \textbf{84.7} & \textbf{89.8} \\
\midrule
\multicolumn{7}{@{}l@{}}{\textbf{VLMs-based methods (RFT)}} \\
AutoVLA~\cite{zhou2025autovla} & 98.4 & 95.6 & \textbf{98.0} & 99.9 & 81.9 & 89.1 \\
ReCogDrive~\cite{li2025recogdrive} & 97.9 & 97.3 & 94.9 & 100.0 & 87.3 & 90.8 \\
AutoDrive-$P^3$~\cite{ye2026autodrivetextp} & 99.1 & 97.4 & 96.5 & 100.0 & 84.8 & 90.6 \\
WAM-Diff~\cite{xu2025wam} & 99.1 & 98.3 & 96.5 & 99.9 & 84.4 & 91.0 \\
SGDrive~\cite{li2026sgdrive} & 98.6 & 97.8 & 96.2 & 100.0 & \textbf{85.8} & 91.1 \\
\midrule
\textbf{DriveStack-VLA (ours)} & \textbf{99.4} & \textbf{98.4} & 96.8 & \textbf{100.0} & 85.7 & \textbf{91.6} \\
\bottomrule
\end{tabular}%
}
\vspace{-12pt} 
\end{table}
\subsubsection{SFT Loss}
We apply the standard autoregressive cross-entropy loss 
$L_{\text{CE}} = -\sum_{n=1}^{|\mathbf{y}|}
\log p_{\theta}\!\left(y_n \mid \mathbf{o}, \mathbf{y}_{<n}\right)$, 
% \begin{equation}
% L_{\text{CE}}(\theta;\mathbf{o},\mathbf{y})
% = -\sum_{n=1}^{|\mathbf{y}|}
% \log p_{\theta}\!\left(y_n \mid \mathbf{o}, \mathbf{y}_{<n}\right),
% \label{eq:ce}
% \end{equation}
% where $\theta$ denotes trainable parameters,
where $p_{\theta}(\cdot)$ is the token distribution from the LLM decoder, $\mathbf{y}=(y_1,\ldots,y_N)$ denotes the full ground-truth target token sequence, $y_n$ denotes token id at position $n$, and $\mathbf{y}_{<n}=(y_1,\ldots,y_{n-1})$ denotes ground-truth prefix used as autoregressive context at step $n$. 
% Finally, let $L^{r}_{\text{CE}}=L_{\text{CE}}(\theta;\mathbf{o}^{r},\mathbf{y})$ denote the loss for the real image pass, and let $L^{m}_{\text{CE}}=L_{\text{CE}}(\theta;\mathbf{o}^{m},\mathbf{y})$ denote the loss for the rasterized image pass with the same targets.
Finally, let $L^{r}_{\text{CE}}=L_{\text{CE}}(\mathbf{o}^{r},\mathbf{y})$ denote the loss for the real image pass, and let $L^{m}_{\text{CE}}=L_{\text{CE}}(\mathbf{o}^{m},\mathbf{y})$ denote the loss for the rasterized image pass with the same targets.
The full Stage-1 loss:
\begin{equation}
L_{\text{SFT}}
=
L^{r}_{\text{CE}}
+\lambda_{\text{meta}}L^{m}_{\text{CE}}
+\lambda_{\text{mask}}L_{\text{mask}}
% +\lambda_{\text{ds}}L_{\text{ds-align}}
+\lambda_{\text{attn}}L_{\text{attn}},
\label{eq:stage1_total}
\end{equation}
% In Equation~\eqref{eq:stage1_total}, $\lambda_{\text{meta}}$, $\lambda_{\text{mask}}$, $\lambda_{\text{ds}}$, and $\lambda_{\text{attn}}$ are scalar weights.
where $\lambda_{\text{meta}}$, $\lambda_{\text{mask}}$, and $\lambda_{\text{attn}}$ are scalar weights.

\subsection{Reinforcement Fine-Tuning} \label{sec:rft}
While Stage-1 trains the actor via token-level supervision, the inference phase relies on best-of-$K$ stochastic sampling. To ensure the sampled proposals achieve high driving rewards and remain strictly decodable, we align the policy using the GRPO algorithm~\cite{shao2024deepseekmath}. Given an observation, the actor samples a candidate group of $K$ action-token sequences $\{\mathbf{a}_i\}_{i=1}^{K}$. The policy is then updated by maximizing the group-relative advantage derived from the candidate rewards. To prevent mode collapse during this update, we introduce a frozen reference model from the Stage-1 checkpoint. This model applies a KL divergence penalty, restricting the actor from drifting excessively from its supervised prior.
% \subsubsection{Reward Formulation}
The advantage is computed based on a joint scalar reward $r_i$ assigned to each candidate. After decoding the token sequence $\mathbf{a}_i$ into a continuous trajectory $\boldsymbol{\tau}_i$, the reward is formulated as:
\begin{equation}
r_i = r_{\text{driving}}(\boldsymbol{\tau}_i) + \alpha_{\text{fmt}} r_{\text{fmt}}(\mathbf{a}_i),
\label{eq:rft_reward}
\end{equation}
where $r_{\text{driving}}(\cdot)$ evaluates each decoded trajectory on three aspects: collision-free to prevent hitting obstacles, collision buffer to maintain safe margins around the ego vehicle, and drivable-area compliance to keep the route within road boundaries. The format reward $r_{\text{fmt}}(\cdot)$ penalizes violations of the action-token specification by enforcing sequence length, index range, and schema constraints. 
% Comprehensive details regarding the formulation of the reward and the objectives of the reinforcement fine-tuning are provided in the supplementary material.

% \subsubsection{GRPO Update.}
% The policy is optimized using GRPO, a critic-free objective for multimodal driving tasks. A group-relative advantage $\hat{A}_i$ is computed by normalizing rewards $\{r_i\}$ within the sampled group. The objective minimizes the following clipped surrogate loss:
% \begin{equation}
% L_{\text{RFT}} = -\frac{1}{K}\sum_{i=1}^{K} \min\!\Big(\rho_i\hat{A}_i, \mathrm{clip}(\rho_i,1-\epsilon_c,1+\epsilon_c)\hat{A}_i\Big) + \beta\,\mathrm{KL}(\pi_{\theta}\,\|\,\pi_{\text{ref}}),
% \label{eq:rft_loss}
% \end{equation}
% where $\rho_i=\pi_{\theta}(\mathbf{a}_i\mid\mathbf{o})/\pi_{\text{old}}(\mathbf{a}_i\mid\mathbf{o})$ is the importance ratio, $\epsilon_c$ is the clipping threshold, and $\beta$ scales the KL penalty to the frozen Stage-1 reference policy $\pi_{\text{ref}}$ to limit drift from initialization. Gradients backpropagate only through the action-token sequence $\mathbf{a}_i$, which calibrates the proposal distribution while preserving overall multimodal behavior.

\subsection{Self-Critic: Scoring and Refinement}
\label{sec:heads}
The critic utilizes the last-layer hidden states of the LLM decoder to enable the scoring and refinement of trajectories. To construct the training data, multiple checkpoints of the actor from various iterations of Stage-1 are evaluated on the training split of the NAVSIM dataset~\cite{dauner2024navsim} to generate diverse candidate trajectories alongside the corresponding Predictive Driver Model Score (PDMS) targets for each observation. To ensure that the critic acquires a robust capability for ranking rather than merely regressing scalar values, approximately $60\%$ of the samples are filtered out; only the observations that contain at least three candidate trajectories and exhibit a distinct gap in PDMS between the highest-scoring and the lowest-scoring candidates are retained.

\subsubsection{Environment Tokens}
Let $\mathbf{H}\in\mathbb{R}^{l\times d}$ be the last-layer hidden states of the LLM decoder for the prefix context, where $l$ is the token length and $d$ is the hidden dimension.
Let $\mathcal{K}_{\text{env}}$ be the union of token positions that correspond to BEV tokens and camera tokens.
We extract environment tokens
$\mathbf{E}=\mathbf{H}[\mathcal{K}_{\text{env}}]\in\mathbb{R}^{l_e\times d}$,
where $l_e=|\mathcal{K}_{\text{env}}|$.
We then project them to a head dimension $d_h$ with a learned linear map $P_h:\mathbb{R}^{d}\rightarrow\mathbb{R}^{d_h}$ and obtain
$\mathbf{E}_h=P_h(\mathbf{E})\in\mathbb{R}^{l_e\times d_h}$ to compute a pooled environment vector $\mathbf{e}_h\in\mathbb{R}^{d_h}$.

\begin{table*}[t]
\caption{Comparison with state-of-the-art methods on the \textbf{NAVSIMv2}~\cite{cao2025pseudo} (navtest). \texttt{False}/\texttt{True} denotes the human penalty filter setting; \texttt{True} disables penalties for cases where the human agent violates constraints, thereby reducing false-positive penalties.}
\vspace{-5pt}
\label{tab:navsim_v2}
\centering
\scriptsize
\setlength{\tabcolsep}{2.3pt}
% \renewcommand{\arraystretch}{0.95}
% 前10列正常，最后两列整列灰底
\begin{tabular*}{\textwidth}{@{\extracolsep{\fill}}@{}lccccccccc>{\columncolor{gray!15}}c>{\columncolor{gray!15}[\tabcolsep][0pt]}c@{}}
\toprule
\multirow{2}{*}{Method} &
\multirow{2}{*}{NC $\uparrow$} &
\multirow{2}{*}{DAC $\uparrow$} &
\multirow{2}{*}{DDC $\uparrow$} &
\multirow{2}{*}{TLC $\uparrow$} &
\multirow{2}{*}{EP $\uparrow$} &
\multirow{2}{*}{TTC $\uparrow$} &
\multirow{2}{*}{LK $\uparrow$} &
\multirow{2}{*}{HC $\uparrow$} &
\multirow{2}{*}{EC $\uparrow$} &
\multicolumn{2}{>{\columncolor{gray!15}}c}{\textbf{EPDMS} $\uparrow$} \\
\cmidrule(lr){11-12}
& & & & & & & & & & \textbf{False} & \textbf{True} \\
\midrule
Human & 100.0 & 100.0 & 99.8 & 100.0 & 87.4 & 100.0 & 100.0 & 98.1 & 90.1 & 90.3 & 94.5 \\
\midrule
TransFuser~\cite{chitta2022transfuser} & 96.9 & 89.9 & 97.8 & 99.7 & 87.1 & 95.4 & 92.7 & 98.3 & 87.2 & 76.7 & 84.0 \\
DiffusionDrive~\cite{liao2025diffusiondrive} & 98.2 & 96.2 & 99.5 & 99.8 & 87.4 & 97.3 & 96.9 & \textbf{98.4} & \textbf{87.7} & 84.7 & 88.2 \\
WoTE~\cite{li2025end} & 98.5 & 96.8 & 98.8 & 99.8 & 86.1 & 97.9 & 95.5 & 98.3 & 82.9 & 84.2 & 87.7 \\
WAM-Diff~\cite{xu2025wam} & 99.0 & 98.4 & 99.3 & 99.9 & 87.0 & 98.6 & 96.2 & 98.1 & 78.5 & -- & 89.7 \\
SGDrive~\cite{li2026sgdrive} & 98.6 & 94.3 & 99.5 & 99.9 & 86.0 & 97.9 & 96.1 & 98.3 & 85.9 & 86.2 & -- \\
AutoDrive-$P^3$~\cite{ye2026autodrivetextp} & 99.1 & 97.4 & 99.2 & 99.8 & \textbf{88.0} & 98.7 & 96.3 & 98.3 &85.5 & 86.2 & 89.9 \\
\midrule
\textbf{DriveStack-VLA} & \textbf{99.4} & \textbf{98.4} & \textbf{99.7} & \textbf{99.9} & 87.4 & \textbf{99.0} & \textbf{98.0} & 97.3 & 82.5 & \textbf{87.3} & \textbf{91.0} \\
\bottomrule
\end{tabular*}
\vspace{-10pt} 
\end{table*}

% \setlength{\tabcolsep}{2.5pt}      % 默认约6pt，调小列间距
% \renewcommand{\arraystretch}{0.95} % 调小行高

% \begin{table*}[tb]
% \caption{Comparison with state-of-the-art methods on the NAVSIMv2~\cite{cao2025pseudo}}
% \label{tab:navsim_v2}
% \centering
% \footnotesize
% \setlength{\tabcolsep}{2.5pt}
% \renewcommand{\arraystretch}{0.95}
% \begin{tabular*}{\textwidth}{@{\extracolsep{\fill}}@{}lccccccccccc@{}}
% \toprule
% Method &
% NC$\uparrow$ & DAC$\uparrow$ & DDC$\uparrow$ & TLC$\uparrow$ & EP$\uparrow$ &
% TTC$\uparrow$ & LK$\uparrow$ & HC$\uparrow$ & EC$\uparrow$ &
% \multicolumn{2}{c}{\textbf{EPDMS}$\uparrow$} \\
% \cmidrule(lr){11-12}
% & & & & & & & & & & False & True \\
% \midrule
% Human & 100.0 & 100.0 & 99.8 & 100.0 & 87.4 & 100.0 & 100.0 & 98.1 & 90.1 & 90.3 & 94.5 \\
% \midrule
% TransFuser~\cite{chitta2022transfuser} & 96.9 & 89.9 & 97.8 & 99.7 & 87.1 & 95.4 & 92.7 & 98.3 & 87.2 & 76.7 & 84.0 \\
% DiffusionDrive~\cite{liao2025diffusiondrive} & 98.2 & 96.2 & 99.5 & 99.8 & 87.4 & 97.3 & 96.9 & 98.4 & 87.7 & 84.7 & 88.2 \\
% WoTE~\cite{li2025end} & 98.5 & 96.8 & 98.8 & 99.8 & 86.1 & 97.9 & 95.5 & 98.3 & 82.9 & 84.2 & 87.7 \\
% WAM-Diff~\cite{xu2025wam} & 99.0 & 98.4 & 99.3 & 99.9 & 87.0 & 98.6 & 96.2 & 98.1 & 78.5 & -- & 89.7 \\
% SGDrive~\cite{li2026sgdrive} & 98.6 & 94.3 & 99.5 & 99.9 & 86.0 & 97.9 & 96.1 & 98.3 & 85.9 & 86.2 & -- \\
% \midrule
% \textbf{DriveStack-VLA (ours)} & 99.4 & 98.4 & 99.7 & 99.9 & 87.4 & 99.0 & 98.0 & 97.3 & 82.5 & 87.3 & 91.0 \\
% \bottomrule
% \end{tabular*}
% \end{table*}
\begin{table}[tb]
\caption{Comparison with state-of-the-art methods on the \textbf{Bench2Drive}~\cite{jia2024bench2drive}.}
\vspace{-5pt}
\label{tab:bench2drive}
\centering
% \footnotesize
\scriptsize
% \begin{tabular*}{\linewidth}{@{\extracolsep{\fill}}@{}lcccc@{}}
\resizebox{\columnwidth}{!}{%
\begin{tabular}{lccc>{\columncolor{gray!15}}c}
\toprule
Method $\uparrow$ & Efficiency $\uparrow$ & Comfortness $\uparrow$ & Success Rate ($\%$) $\uparrow$ & \textbf{Driving Score} $\uparrow$ \\
\midrule
AD-MLP~\cite{zhai2023rethinking} & 48.45 & 22.63 & 0.00 & 18.05 \\
UniAD-Base~\cite{hu2023planning} & 129.21 & 43.58 & 16.36 & 45.81 \\
VAD~\cite{jiang2023vad} & 157.94 & \textbf{46.01} & 15.00 & 42.35 \\
RAP~\cite{feng2025rap} & \textbf{165.47} & 23.63 & 37.27 & 66.42 \\
ReCogDrive~\cite{li2025recogdrive} & 138.18 & 17.45 & 45.45 & 71.36 \\
Orion~\cite{fu2025orion} & 151.48 & 17.38 & 54.62 & 77.74 \\
AutoVLA~\cite{zhou2025autovla} & 146.93 & 39.33 & \textbf{57.73} & 78.84 \\
\midrule
\textbf{DriveStack-VLA} & 164.52 & 11.31 & 56.36 & \textbf{79.49} \\
\bottomrule
\end{tabular}%
}
\vspace{-10pt} 
\end{table}
\subsubsection{Scoring Head}
Given a candidate trajectory $\boldsymbol{\tau}$ and the pooled environment vector $\mathbf{e}_h$, a scoring head $S(\boldsymbol{\tau},\mathbf{e}_h)$ outputs a normalized predicted score $\hat{s}\in[0,1]$.
We train the scoring head with a regression loss and a ranking loss.
Let $\tilde{s}\in[0,1]$ denote the target score for a candidate, and let $(b,w)$ denote the indices of the best and worst candidates within the same scene according to the target scores.
We define a gap weight $g=\mathrm{clip}(\tilde{s}_b-\tilde{s}_w,0,1)$.
The score loss is
\begin{equation}
L_{\text{score}}
=
\mathrm{SmoothL1}(\hat{s}-\tilde{s})
-\lambda_{\text{rank}}\,g\,
\log\sigma\!\left(\frac{\hat{s}_{b}-\hat{s}_{w}}{T_r}\right),
\label{eq:score_loss}
\end{equation}
 where $\mathrm{SmoothL1}(\cdot)$ is the SmoothL1 (Huber) loss, $\sigma(z)=1/(1+\exp(-z))$ is the logistic sigmoid, $T_r>0$ denotes a temperature that controls the scale of the ranking term, and $\lambda_{\text{rank}}\geq 0$ is a weight.

\subsubsection{Refinement Head.}
The refinement head $R(\boldsymbol{\tau}_0,\mathbf{E}_h)$ processes an initial trajectory $\boldsymbol{\tau}_0$ alongside environment tokens $\mathbf{E}_h$ to predict a bounded residual, thereby generating a refined trajectory:
\begin{equation}
\boldsymbol{\tau}_{\text{ref}}
=
\boldsymbol{\tau}_0
+
\Delta_{\max}\odot\tanh\!\big(R(\boldsymbol{\tau}_0,\mathbf{E}_h)\big),
\label{eq:refine}
\end{equation}
where $\Delta_{\max}$ represents a dimension-wise bound sharing the exact shape of $\boldsymbol{\tau}_0$, $\tanh(\cdot)$ is applied element-wise, and $\odot$ denotes element-wise multiplication.
During training, the second-best trajectory candidate within a scene is selected as the initial trajectory $\boldsymbol{\tau}_0$, while the best candidate serves as the supervision target $\boldsymbol{\tilde{\tau}}$. The network is optimized utilizing a SmoothL1 loss function, formulated as $\mathrm{SmoothL1}(\boldsymbol{\tau}_{\text{ref}},\boldsymbol{\tilde{\tau}})$.
% \begin{equation}
% L_{\text{ref}}
% =
% \mathrm{SmoothL1}(\boldsymbol{\tau}_{\text{ref}},\boldsymbol{\tilde{\tau}}).
% \label{eq:ref_loss}
% \end{equation}

% Finally, the scoring head and the refinement head are trained in separate training runs.

\section{Experiments}
\label{sec:exp}

\subsection{Experimental Setup}
\subsubsection{Datasets and Metrics}
We train the DriveStack-VLA model on both real-world and simulated datasets. For real-world data, the Stage-1 and Stage-2 training processes utilize the NAVSIM dataset~\cite{dauner2024navsim}, which contains 102k training samples. During Stage-1, we construct rasterized images 
% utilizing ground-truth annotations of current-frame map primitives, agent bounding boxes, and traffic light states 
following RAP~\cite{feng2025rap}. The Stage-3 training follows \cref{sec:heads} and utilizes a filtered subset of 40k samples from the NAVSIM dataset. For simulated data, we initialize the model from the checkpoint pretrained on real-world data and perform supervised fine-tuning on the base set of Bench2Drive~\cite{jia2024bench2drive}, which consists of 1,000 clips (950 clips for training and 50 clips for open-loop validation).

% We evaluate the model on two benchmarks. We evaluate open-loop planning on NAVSIM-v1~\cite{dauner2024navsim} using the official PDMS metric, which aggregates multiple planning-related sub-metrics including \emph{No-at-fault Collisions} (NC), \emph{Drivable Area Compliance} (DAC), \emph{Time-to-Collision within bound} (TTC), \emph{Comfort} (Comf.), and \emph{Ego Progress} (EP). NAVSIM-v2~\cite{cao2025pseudo} extends the evaluation with additional rule checks (e.g., traffic light compliance and lane keeping) and reports Extended Predictive Driver Model Score (EPDMS). We report results under both settings of \texttt{human\_penalty\_filter} (False/True) to match prior work. Bench2Drive evaluates end-to-end driving in closed-loop simulation with reactive agents. The closed-loop evaluation covers 220 routes across CARLA~\cite{dosovitskiy2017carla} towns, where each route contains a safety-critical event, and reports four metrics: Success Rate, Driving Score, Efficiency, and Comfortness.

% We evaluate DriveStack-VLA on two benchmarks. Rasterized images highlight safety-critical elements in a controllable manner. Following the methodology of RAP~\cite{feng2025rap}, we construct these rasterized images utilizing ground-truth annotations of current-frame map primitives, agent bounding boxes, and traffic light states. We subsequently employ these rasterized images as a teacher modality to align the perceptual focus of the real images with that of the renderings through masked camera-token alignment and action-to-vision attention distillation~\cite{zagoruyko2016paying}.
We evaluate open-loop planning on NAVSIMv1~\cite{dauner2024navsim} using the official PDMS metric, which aggregates multiple planning-related sub-metrics including No-at-fault Collisions (NC), Drivable Area Compliance (DAC), Time-to-Collision within bound (TTC), Comfort (Comf.), and Ego Progress (EP). NAVSIMv2~\cite{cao2025pseudo} proposes a two-stage pseudo-simulation evaluation protocol. The core metric is the Extended Predictive Driver Model Score (EPDMS), which extends PDMS with additional factors including Driving Direction Compliance (DDC), Traffic Light Compliance (TLC), Lane Keeping (LK), History Comfort (HC), and Extended Comfort (EC). We report results under both settings of the human penalty filter (\texttt{False}/\texttt{True}). The closed-loop evaluation covers 220 routes across CARLA~\cite{dosovitskiy2017carla} towns, where each route contains a safety-critical event, and reports four metrics: Driving Score, Success Rate, Efficiency, and Comfortness.
%Navhard
 % for traffic-rule compliance and temporal consistency, 
% The second stage further synthesizes counterfactual camera observations after policy deviations to approximate closed-loop feedback from logged data. 

% \subsubsection{Implementation Details.}
%All experiments use the same action representation described in Sec.~\ref{sec:method}, with a 4-second planning horizon at 2 Hz. 图像被调整为保持原始纵横比，但减少到 32 × 32 × 384 像素. All three training stages are completed on 4*8 NVIDIA A100 GPUs。 stage1训练了4epochs with AdamW  learning rate $1\times10^{-4}$, a cosine learning-rate schedule, a warm-up ratio of 0.03 . Because each training sample is expanded into a paired (real, rasterized) forward pass, the effective micro-batch is 2. For Render-Teacher Alignment, we set $\tau=0.2$ and $\gamma=2$ for the render-guided mask。In sft loss function we set $\lambda_{\text{meta}}=1.0$, $\lambda_{\text{mask}}=0.02$, and $\lambda_{\text{attn}}=0.02$ 。 stage2 训练了 3000 steps ， sample a group of \textbf{$K=3$} candidate trajectories per observation, use clipping range 0.18, and a KL penalty coefficient $\beta=0.02$ ， use a learning rate of $2\times10^{-6}$。 Stage3 分别为score head和refinement head训练了10epochs， use AdamW with learning rate $3\cdot10^{-4}$. The score head uses SmoothL1 regression plus a ranking loss with $\lambda_{\text{rank}}=1.0$ and $T_r=0.5$。The refinement head predicts a bounded residual with $\Delta_{\max}=(3\text{m},3\text{m},0.3\text{rad})$ and is trained with SmoothL1 supervision.

% \input{figures/fig4_trajvis}

% \begin{table}[tb]
% \begin{table}[htbp]
\begin{table*}[t]
\caption{Results on the Navhard benchmark. * denotes results reproduced with the official code repository or official checkpoint.}
\vspace{-5pt}
\label{tab:navhard_epdms}
\centering
\scriptsize
% \begin{tabular*}{\textwidth}{@{\extracolsep{\fill}}@{}lcccccccccc>{\columncolor{gray!15}[\tabcolsep][0pt]}c@{}}
\begin{tabular*}{\textwidth}{@{\extracolsep{\fill}}@{}lcccccccccc>{\columncolor{gray!15}[\tabcolsep][1.5pt]}c@{}}
\toprule
Method & Stage & NC$\uparrow$ & DAC$\uparrow$ & DDC$\uparrow$ & TLC$\uparrow$ & EP$\uparrow$ & TTC$\uparrow$ & LK$\uparrow$ & HC$\uparrow$ & EC$\uparrow$ & \textbf{EPDMS}$\uparrow$ \\
\midrule
\multirow{2}{*}{TransFuser*~\cite{chitta2022transfuser}} 
 & S1 & 96.2 & 79.5 & \textbf{99.1} & 99.5 & 84.1 & 95.1 & 94.2 & \textbf{97.5} & \textbf{79.1} & \\
 & S2 & 77.7 & 70.2 & 84.2 & 98.0 & 85.1 & 75.6 & 45.4 & \textbf{95.7} & \textbf{75.9} & \multirow{-2}{*}{23.1} \\
\midrule
\multirow{2}{*}{DiffusionDrive*~\cite{liao2025diffusiondrive}} 
 & S1 & 96.0 & 79.7 & 97.4 & 99.5 & 81.3 & 93.1 & 90.8 & 96.8 & 73.8 & \\
 & S2 & \textbf{82.1} & 72.2 & \textbf{88.5} & \textbf{98.7} & 85.1 & \textbf{78.8} & 49.2 & 89.3 & 71.2 & \multirow{-2}{*}{24.2} \\
\midrule
\multirow{2}{*}{\textbf{DriveStack-VLA (ours)}}
 & S1 & \textbf{97.9} & \textbf{91.6} & 98.7 & \textbf{99.8} & \textbf{85.2} & \textbf{96.4} & \textbf{96.4} & 97.3 & 63.6 & \\
 & S2 & 81.0 & \textbf{78.2} & 86.5 & 97.9 & \textbf{86.1} & 76.7 & \textbf{49.6} & \textbf{95.7} & 50.2 & \multirow{-2}{*}{\textbf{34.9}} \\
\bottomrule
\end{tabular*}
\vspace{-10pt}
\end{table*}

\begin{figure*}[t]
      \centering
       % 1. 压减图片与 Caption 之间的距离
      \setlength{\abovecaptionskip}{4pt} 
      % 2. 压减 Caption 与下方正文之间的距离
      \setlength{\belowcaptionskip}{-12pt}
      \includegraphics[width=\linewidth]{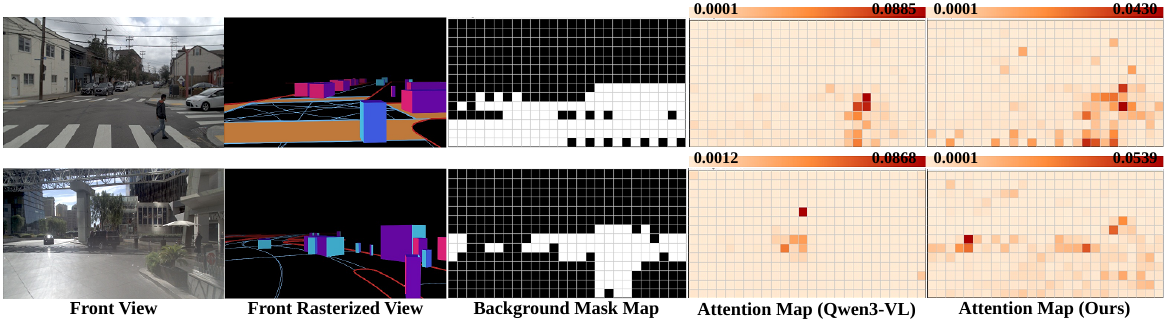}
      % \vspace{-1pt}
      \captionsetup{font={small}}
      % \caption{}
      \caption{Qualitative analysis of Render-Teacher Alignment.}
      \label{fig:attnmap}
\end{figure*}

\subsubsection{Implementation Details}
% All experiments use the action representation described in \cref{sec:method}, with a 4-second planning horizon at 2 Hz. Images are resized to maintain original aspect ratios but reduced to $32\times 32\times 384$ visual pixels. All three training stages are completed on $4\times 8$ NVIDIA A100 GPUs.
% Stage-1 trains for 4 epochs with AdamW, a learning rate of $1\times 10^{-4}$, a cosine learning-rate schedule, and a warm-up ratio of 0.03. Because each training sample is expanded into a paired (real, rasterized) forward pass, the per-GPU batch size is 2. For Render-Teacher Alignment, we set $\tau=0.2$ and $\gamma=2$ for the mask. In the SFT loss, we set $\lambda_{\text{meta}}=1.0$, $\lambda_{\text{mask}}=0.02$, and $\lambda_{\text{attn}}=0.02$.
% Stage-2 runs for 3,000 steps, samples a group of $K=3$ candidate trajectories per observation, uses a clipping range of 0.18, and applies a KL penalty coefficient of $\beta=0.02$. The learning rate is set to $2\times 10^{-6}$. 
% Stage-3 trains the scoring head and the refinement head for 10 epochs each, using AdamW with a learning rate of $3\times 10^{-4}$. The scoring head uses SmoothL1 regression with an additional ranking loss, where $\lambda_{\text{rank}}=1.0$ and $T_r=0.5$. The refinement head predicts a bounded residual with $\Delta_{\max}=(3\text{m},3\text{m},0.3\text{rad})$ and is trained with SmoothL1 supervision.
All experiments utilize the action representation detailed in \cref{sec:method}, which features a 4-second planning horizon at 2 Hz. The input images are resized to preserve the original aspect ratios while being reduced to a resolution of $32\times 32\times 384$ visual pixels. The three stages of training are executed on a cluster of 32 NVIDIA A100 GPUs. During Stage-1, the model is trained for 4 epochs utilizing the AdamW optimizer with a cosine learning-rate schedule. To accommodate the paired forward pass of real and rasterized images, the batch size per GPU is set to 2. In Stage-2, the RFT process is executed for 1 epoch. For Stage-3, the scoring head and the refinement head are trained for 10 epochs each to predict the score of the trajectory and the bounded residuals. 

% Comprehensive details regarding the hyperparameter configurations for all training stages are provided in the supplementary material.

\subsection{Main Results}
\subsubsection{Benchmark Results}
%\Cref{tab:navsim_v1} shows the NAVSIM v1 benchmark, where DriveStack-VLA achieves the highest overall PDMS score of 89.8 与 91.6 分别在SFT和RFT训练策略中，surpassing all prior VLMs-based methods. Notably, 与我们的vlm backbone Qwen3-VL-4B微调后相比， 在SFT训练策略下，我们通过训练stage-1的sft过程与stage-3的score and refinement heads，带来了 +5.8 的PDMS提升 （improves PDMS by +5.8 over our the vlm backbone Qwen3-VL-4B）。在包含了RFT训练策略下，DriveStack-VLA完成了三阶段的训练流程，实现了 91.6 PDMS，并且在NC和DAC两个safety-critical关键指标上实现了最强结果。\Cref{tab:navsim_v2} reports EPDMS on NAVSIMv2 under both settings of the human penalty filter. DriveStack-VLA achieves 91.0 EPDMS when the filter is enabled and 87.3 when it is disabled also sets a new state-of-the-art。
\Cref{tab:navsim_v1} reports results on the NAVSIMv1 benchmark, where Drive-Stack-VLA attains overall PDMS scores of \textbf{89.8} and \textbf{91.6} under the SFT and RFT training strategies, respectively, outperforming all prior VLM-based methods. Beyond the strongest model, proposed method yields consistent improvements over a baseline that directly fine-tunes the Qwen3-VL-4B backbone under identical training configurations. Specifically, the integration of the Stage-1 SFT and the Stage-3 scoring and refinement heads increases the PDMS by \textbf{+5.8}, while the addition of the Stage-2 RFT further elevates the PDMS to 91.6. The complete model obtains the strongest results on the safety-critical metrics, namely NC and DAC. \Cref{tab:navsim_v2} reports EPDMS on NAVSIM-v2 under both settings of the human penalty filter. DriveStack-VLA achieves \textbf{91.0} EPDMS when the filter is enabled and \textbf{87.3} when the filter is disabled, establishing a new state of the art under both evaluation settings.

% \subsubsection{Bench2Drive.}
% As show in \cref{tab:bench2drive},DriveStack-VLA achieves the best overall performance reaches the critic metric top driving score of \textbf{79.49} and the highest Success Rate (56.36\%)  These results highlight the effectiveness and reliability of our framework in complex urban yields robust closed-loop behavior.
% As shown in \cref{tab:bench2drive}, DriveStack-VLA achieves strong closed-loop performance on Bench2Drive, with a driving score of \textbf{79.49} and a success rate of \textbf{56.36\%}. These results indicate that the proposed framework yields robust behavior in complex urban simulation with reactive agents.
As shown in \cref{tab:bench2drive}, DriveStack-VLA achieves a competitive driving score of \textbf{79.49} and a success rate of \textbf{56.36\%} on Bench2Drive. Notably, this performance is attained via standard supervised fine-tuning of the real-world pretrained model, without utilizing RFT. This demonstrates the strong transferability and robust behavior of the base architecture in complex urban simulations.

We further evaluate DriveStack-VLA on the more challenging Navhard benchmark, which employs Gaussian splatting to generate scenarios beyond the training data distribution and adopts a two-stage evaluation protocol. As demonstrated in \cref{tab:navhard_epdms}, DriveStack-VLA achieves an overall EPDMS of 34.9, maintaining robust performance across both the S1 evaluation and the more demanding S2 counterfactual evaluation.

\subsubsection{Qualitative results}
% As illustrated in \cref{fig:trajvis}, we compare the trajectories of the proposed model (Stage-1 and Stage-3, excluding the Stage-2 RFT process) with the baseline Qwen3-VL~\cite{bai2025qwen3} under two challenging scenarios: a narrow intersection with dense traffic and a route with intense sunlight reflection. While the baseline model suffers from severe deviations and potential risks of collision due to degraded perceptual cues, the proposed Stage-1 model consistently generates safe, collision-free trajectories. Furthermore, the Stage-3 module provides effective residual refinement to further enhance the overall quality of planning.

To analyze the effectiveness of the alignment method described in \cref{sec:rgm3}, \cref{fig:attnmap} visualizes the action-to-vision attention heat maps alongside the rasterized images and background masks. The attention values are extracted from the last layer of the LLM decoder and mapped back to the spatial patches of the original image grid (tokenized into $14\times 26$). Consequently, brighter regions indicate a stronger influence on the generation of actions. In the first scenario, the baseline model focuses exclusively on the foreground pedestrian and fails to attend to the vehicles at the side of the intersection. Conversely, the proposed model places strong attention on all safety-critical agents. In the second scenario, the baseline model is easily distracted by high-intensity background areas caused by strong sunlight reflection, whereas the proposed model successfully highlights the lane structure and nearby vehicles. These qualitative results validate that the masked camera-token alignment and the action-to-vision attention distillation enable the model to suppress irrelevant backgrounds and concentrate on planning-relevant visual evidence under challenging conditions. More qualitative visualizations are available on the project page.

\begin{table}[tb]
% \captionsetup{justification=raggedright,singlelinecheck=false}
\caption{Ablation on BEV DeepStack and Rasterized Image Feature Alignment.}
\vspace{-5pt}
\label{tab:ablate_bev_render}
\centering
\scriptsize
\setlength{\tabcolsep}{1.8pt}
\renewcommand{\arraystretch}{0.9}
\resizebox{\columnwidth}{!}{%
\begin{tabular}{@{}ccccccccccc>{\columncolor{gray!15}}c@{}}
\toprule
Exp. & BEV & DeepBEV & RI & UMMSE & MMSE & AttnDistill &
NC$\uparrow$ & DAC$\uparrow$ & TTC$\uparrow$ & EP$\uparrow$ & \textbf{PDMS}$\uparrow$ \\
\midrule
a & $\times$ & $\times$ & $\times$ & $\times$ & $\times$ & $\times$ & 97.7 & 92.4 & 93.9 & 78.5 & 84.0 \\
b & $\checkmark$ & $\times$ & $\times$ & $\times$ & $\times$ & $\times$ & 97.8 & 93.7 & 94.0 & 79.5 & 85.1 \\
c & $\checkmark$ & $\checkmark$ & $\times$ & $\times$ & $\times$ & $\times$ & 98.1 & 94.9 & 94.8 & 80.2 & 86.6 \\
d & $\checkmark$ & $\checkmark$ & $\checkmark$ & $\times$ & $\times$ & $\times$ & 97.9 & 94.6 & 94.7 & 80.6 & 86.4 \\
e & $\checkmark$ & $\checkmark$ & $\checkmark$ & $\checkmark$ & $\times$ & $\times$ & 98.0 & 95.3 & 94.4 & 81.3 & 87.0 \\
f & $\checkmark$ & $\checkmark$ & $\checkmark$ & $\times$ & $\checkmark$ & $\times$ & 98.2 & 95.6 & 94.8 & 81.7 & 87.5 \\
g & $\checkmark$ & $\checkmark$ & $\checkmark$ & $\times$ & $\checkmark$ & $\checkmark$ & \textbf{98.2} & \textbf{96.3} & \textbf{95.0} & \textbf{82.4} & \textbf{88.2} \\
\bottomrule
\end{tabular}%
}
\vspace{-10pt}
\end{table}

\begin{table}[tb]
\caption{Ablation on Self-critic and RFT Reward.}
\vspace{-5pt}
\label{tab:ablate_critic_rft}
\centering
\tiny
\setlength{\tabcolsep}{1.2pt}

\resizebox{\columnwidth}{!}{%
\begin{tabular}{cccccccc>{\columncolor{gray!15}}c}
\toprule
% Exp & BEV & BEV DeepStack & Render & CE Loss & Unmasked MSE & Masked MSE & Attn Distill &
% NC$\uparrow$ & DAC$\uparrow$ & TTC$\uparrow$ & EP$\uparrow$ & \textbf{PDMS}$\uparrow$ \\
Exp. & Self-critic & Format Reward & Driving Reward & NC$\uparrow$ & DAC$\uparrow$ & TTC$\uparrow$ & EP$\uparrow$ & \textbf{PDMS}$\uparrow$ \\
\midrule
a & $\times$ & $\times$ & $\times$ & 98.2 & 96.3 & 95.0 & 82.4 & 88.2 \\
% b & $\checkmark$ & $\times$ & $\times$ & $\times$ & 98.8 & 96.9 & 94.7 & 84.3 & 89.3 \\
b & $\checkmark$  & $\times$ & $\times$ & 98.8 & 97.2 & 94.9 & \textbf{84.7} & 89.8 \\
c & $\times$ & $\checkmark$ & $\times$ & 98.7 & 97.0 & 94.7 & 84.5 & 89.5 \\
d & $\times$ & $\checkmark$ & $\checkmark$ & \textbf{99.2} & \textbf{97.6} & \textbf{96.9} & 84.2 & \textbf{90.5} \\
\bottomrule
\end{tabular}%
}
\vspace{-12pt} 
\end{table}

\subsection{Ablation Studies}
% All ablations are conducted on NAVSIM-v1 and report NC/DAC/TTC/EP/PDMS in $[0,100]$.

\subsubsection{BEV DeepStack and Render-Teacher Alignment}
% 我们在stage-1阶段训练中评估了BEV DeepStack与Render-Teacher Alignment（换成和前面intro时的说法）的有效性，如 \cref{tab:ablate_bev_render} 所示，表格中BEV代表是否有BEV注入仅将（BEV作为额外的token），DeepBEV代表的是进行Deepstack注入的BEV，RI代表训练时是否加入Rasterized images（没有额外的对齐loss），UMMSE代表无masked的mse loss进行alignment，MMSE代表 masked mse，attnDistill代表是否进行action-to-vision attention distillation. 如Exp.(c)所示，与Exp.(a)相比deepstack的bev注入能带来 +2.5的PDMS提升，并且比仅把BEV当作额外的token注入更有效。 如Exp.(d)所示，简单的添加rasterized images 并不能带来收益，而如Exp.(e)所示，不进行mask直接进行mse的token级别特征对齐，也只能带来有限的改进。如Exp.(f)和Exp.(g)所示，与Exp(d)相比，我们提出的render-teacher alignment方法显著提高了PDMS +2.2，值得注意的是单独DAC这项指标就提升了2.3，说明它显著增强了action token在规划过程中关注安全关键的视觉区域。
We evaluate the effectiveness of BEV DeepStack Injection and Render-Teacher Alignment in Stage-1 training. As shown in \cref{tab:ablate_bev_render}, \emph{BEV} indicates whether BEV tokens are provided as additional visual tokens, whereas \emph{DeepBEV} denotes BEV injection through the DeepStack connection. \emph{RI} indicates whether rasterized images are included during training without additional alignment losses. \emph{UMMSE} and \emph{MMSE} correspond to unmasked and masked camera-token MSE alignment losses, respectively, and \emph{attnDistill} indicates whether action-to-vision attention distillation is applied. As shown by Exp.(c), DeepStack-based BEV injection improves PDMS by 2.5 compared with Exp.(a) and is more effective than treating BEV as a simple additional token stream. Exp.(d) shows that naively adding rasterized images without alignment does not yield measurable gains. Exp.(e) indicates that unmasked token-level MSE alignment provides only limited improvement, which is consistent with the dominance of background regions in rasterized views. In contrast, Exp.(f) and Exp.(g) show that the proposed Render-Teacher Alignment significantly improves PDMS by 2.2 over Exp.(d). A notable gain is observed on DAC, which improves by 2.3, suggesting that the alignment encourages action tokens to attend to safety-critical visual regions during planning.

%bev
% \Cref{tab:ablate_bev_render} confirms that introducing BEV cues consistently improves planning.
% A plain BEV branch yields only a marginal gain over the camera-only baseline, whereas injecting BEV tokens through the DeepStack multi-level pathway provides a stronger and more stable geometric prior, improving PDMS and drivable-area compliance without increasing the autoregressive context length substantially.

%render
% shows that naive render augmentation (adding rasterized views but training only with CE) yields limited improvement.
% Direct alignment without masking is hindered by the large black background regions in rasterized views.
% Our full Render-Teacher Alignment (masked camera-token MSE + action-to-vision attention distillation) improves both DAC and EP, indicating that it not only transfers appearance cues but also encourages action 
% tokens to attend to safety-critical visual regions during planning.

\subsubsection{Self-Critic and RFT Reward}
We further investigate the contributions of the self-critic module and the RFT objectives built upon the Stage-1 model. As presented in \cref{tab:ablate_critic_rft}, Exp.(b) improves PDMS by 1.6 over Exp.(a), indicating that the self-critic can rank and refine actor-generated trajectories without an oracle scorer. Exp.(c) and Exp.(d) further show the benefits of RFT, validating the format and driving rewards for sampling-based planning.

% Comprehensive evaluations of inference latency and additional ablation studies concerning hyperparameter configurations are provided in the supplementary material.

% \Cref{tab:ablate_critic_rft} demonstrates that the score head improves best-of-$N$ selection without requiring any oracle scorer.
% Moreover, the gated refinement head provides additional gains, particularly on ego progress and collision-related metrics, by applying bounded residual corrections only when the selected candidate is uncertain.
% Increasing $N$ yields consistent improvements, and we use as the default for NAVSIMv1 in our full system.

\section{Conclusions}
% We propose DriveStack-VLA, an end-to-end driving framework coupling a multimodal large language model with an enhanced visual stack. It integrates BEV DeepStack injection and Render-Teacher Alignment to establish geometric priors and emphasize safety-critical cues. Additionally, reinforcement fine-tuning and a lightweight self-critic module align the proposal distribution, enabling efficient trajectory ranking and refinement without additional text generation. Experiments demonstrate state-of-the-art performance on open-loop and closed-loop planning benchmarks.
We propose DriveStack-VLA, an end-to-end driving framework coupling a multimodal large language model with an enhanced visual stack. It integrates BEV DeepStack injection and Render-Teacher Alignment to establish geometric priors and emphasize safety-critical cues. Experiments demonstrate state-of-the-art performance on open-loop and closed-loop planning benchmarks.

% Furthermore, DriveStack-VLA achieves an exceptional balance between the performance of planning and the efficiency of inference. Even with the integration of multi-trajectory sampling, scoring, and refinement, the best-performing model reported in this study maintains a total inference latency of approximately 600 milliseconds. Notably, the efficiency of the proposed framework surpasses that of previous end-to-end VLA models relying on an action codebook or text-based waypoints.
% Integrated supplementary material for the IEEE main document.
\clearpage
\begingroup

\section*{Supplementary Material}

\setcounter{section}{0}
\setcounter{subsection}{0}
\setcounter{subsubsection}{0}
\setcounter{figure}{0}
\setcounter{table}{0}
\setcounter{equation}{0}

\renewcommand{\thesection}{S\arabic{section}}
\renewcommand{\thesubsection}{\thesection-\Alph{subsection}}
\renewcommand{\thesubsubsection}{\thesubsection.\arabic{subsubsection}}
\renewcommand{\thesectiondis}{\thesection.}
\renewcommand{\thesubsectiondis}{\Alph{subsection}.}
\renewcommand{\thesubsubsectiondis}{\arabic{subsubsection})}
\renewcommand{\thefigure}{S\arabic{figure}}
\renewcommand{\thetable}{S\arabic{table}}
\renewcommand{\theequation}{S\arabic{equation}}

\makeatletter
\@ifundefined{theHsection}{}{\renewcommand{\theHsection}{supp.\arabic{section}}}
\@ifundefined{theHsubsection}{}{\renewcommand{\theHsubsection}{supp.\arabic{section}.\arabic{subsection}}}
\@ifundefined{theHsubsubsection}{}{\renewcommand{\theHsubsubsection}{supp.\arabic{section}.\arabic{subsection}.\arabic{subsubsection}}}
\@ifundefined{theHfigure}{}{\renewcommand{\theHfigure}{supp.fig.\arabic{figure}}}
\@ifundefined{theHtable}{}{\renewcommand{\theHtable}{supp.tab.\arabic{table}}}
\@ifundefined{theHequation}{}{\renewcommand{\theHequation}{supp.eq.\arabic{equation}}}
\@ifundefined{c@algocf}{}{\setcounter{algocf}{0}}
\@ifundefined{thealgocf}{}{\renewcommand{\thealgocf}{S\arabic{algocf}}}
\@ifundefined{theHalgocf}{}{\renewcommand{\theHalgocf}{supp.alg.\arabic{algocf}}}
\makeatother

\SetAlgoNoLine
\DontPrintSemicolon
\SetAlgoCaptionSeparator{ }
\SetNlSty{}{}{:}

\section{Details of Reinforcement Fine-Tuning}
\label{sec:details_of_rft}
This section supplements Stage-2 (RFT) by specifying the GRPO objective, the KL regularization estimator, the joint reward design, the candidate sampling protocol, and the key implementation details used for optimization.

\subsection{KL regularization against a frozen reference policy}
\label{supp:sec_kl_regularization}
During Stage-2, RFT aligns the proposal distribution for sampling-based driving
with a GRPO objective. Given an observation $o=(x,I,u)$, where $x$ is the
navigation instruction, $I$ is the set of multi-view images, and $u$ is the
ego state, the actor samples a group of $K$ candidate action-token sequences
$\{\mathbf{a}_i\}_{i=1}^{K}$. To avoid mode collapse during policy updates and
to keep the actor from moving too far away from the supervised prior learned
in Stage-1, we initialize a frozen reference policy
$\pi_{\mathrm{ref}}$ from the Stage-1 checkpoint and apply the following
sample-based KL penalty to each sampled candidate:
\begin{equation}
\widehat{D}_{\mathrm{KL},i}
=
\frac{\pi_{\mathrm{ref}}(\mathbf{a}_i \mid o)}{\pi_{\theta}(\mathbf{a}_i \mid o)}
-
\log\!\left(
\frac{\pi_{\mathrm{ref}}(\mathbf{a}_i \mid o)}{\pi_{\theta}(\mathbf{a}_i \mid o)}
\right)
- 1,
\label{supp:eq_kl_est}
\end{equation}
where the probability ratio is computed from the corresponding sequence
log-likelihoods 
$\log \pi(\mathbf{a}_i \mid o)=\sum_{t}\log \pi(a_{i,t}\mid o,\mathbf{a}_{i,<t})$.
This penalty is zero when the current policy and the reference policy assign
the same probability to the sampled candidate, and it increases as the gap
between them grows. In this way, it stabilizes GRPO updates, keeps the actor
close to the Stage-1 policy, and helps maintain a stable proposal distribution
under best-of-$K$ sampling.

\subsection{Reward design in Stage-2 reinforcement fine-tuning}
\label{supp:sec_rft_reward}
% Building on the action tokenization in Sec.~3.1 and the Stage-2 reward
% formulation in Sec.~3.4 of the main text, we here specify the concrete
% implementations of the driving reward and the format-consistency reward used in
% reinforcement fine-tuning, together with the corresponding hyperparameter
% settings. For a sampled candidate action-token sequence \(\mathbf{a}_i\), the
% same frozen action codebook as in the main text decodes a valid sequence into a
% continuous trajectory \(\boldsymbol{\tau}_i\). Candidates that violate the
% action-token grammar are treated as invalid proposals, assigned zero driving
% reward, and optimized through the format-consistency term only.
Building upon the established action tokenization and the Stage-2 reward formulation, we detail the concrete implementations of the driving reward and the format-consistency reward utilized in reinforcement fine-tuning, alongside the corresponding hyperparameter settings. For a sampled candidate action-token sequence \(\mathbf{a}_i\), a frozen action codebook decodes the valid sequence into a continuous trajectory \(\boldsymbol{\tau}_i\). Candidates violating the action-token grammar are treated as invalid proposals, assigned a driving reward of zero, and optimized exclusively through the format-consistency term.

\subsubsection{Driving reward.}
% Consistent with Sec.~3.4 of the main text, we define the Stage-2 driving reward
% using a simplified subset of planning signals from the NAVSIMv1~\cite{dauner2024navsim} evaluation
% protocol. Specifically, we use NC, TTC, and DAC, the three sub-metrics most
% directly tied to collision avoidance, safety margin, and drivable-area
% compliance. Although the official PDMS metric captures broader aspects of
% driving quality, including Comfort and Ego Progress, the Stage-2 design focuses
% on the signals most directly related to safety-critical feasibility during
% proposal generation. This provides a lightweight but effective proxy for
% proposal-level training, whose role is to steer the candidate distribution
% toward safe, feasible, and strictly decodable multi-trajectory proposals under best-of-\(K\) sampling. 
We define the driving reward in Stage-2 using a simplified subset of planning signals derived from the evaluation protocol of NAVSIMv1~\cite{dauner2024navsim}. Specifically, we employ No Collision (NC), Time-to-Collision (TTC), and Drivable Area Compliance (DAC), which are the three sub-metrics most directly associated with collision avoidance, collision buffers, and drivable-area compliance. Although the official PDMS metric captures broader aspects of driving quality, including kinematic comfort and route progression, this reward design focuses on the signals most directly related to safety-critical feasibility during the generation of proposals. This formulation provides a lightweight but effective proxy for proposal-level training, which steers the distribution of candidates toward safe, feasible, and strictly decodable multi-trajectory proposals under best-of-\(K\) sampling. 
% Broader aspects of trajectory quality, including the trade-offs reflected by PDMS, are handled later by the Stage-3 scoring and refinement heads.

In practice, we reuse the corresponding NAVSIMv1 planning sub-scores and
combine them into a simple aggregate reward. Let
\(\mathrm{NC}(\boldsymbol{\tau}_i)\), \(\mathrm{TTC}(\boldsymbol{\tau}_i)\),
and \(\mathrm{DAC}(\boldsymbol{\tau}_i)\) denote the corresponding trajectory
evaluation scores, and let
\(s_{\mathrm{NC}}(\boldsymbol{\tau}_i)\),
\(s_{\mathrm{TTC}}(\boldsymbol{\tau}_i)\), and
\(s_{\mathrm{DAC}}(\boldsymbol{\tau}_i)\) denote the normalized versions. We
then define the driving reward as the average of these three sub-scores:
\begin{equation}
r_{\mathrm{driving}}(\boldsymbol{\tau}_i)
=
\frac{
s_{\mathrm{NC}}(\boldsymbol{\tau}_i)
+
s_{\mathrm{TTC}}(\boldsymbol{\tau}_i)
+
s_{\mathrm{DAC}}(\boldsymbol{\tau}_i)
}{3}.
\label{supp:eq_reward_driving}
\end{equation}

\subsubsection{Format reward.}
Complementary to the trajectory-space driving term, the format reward operates
in token space to preserve strict decodability under the action-token
specification. We define
\begin{equation}
\begin{aligned}
r_{\mathrm{fmt}}(\mathbf{a}_i)
&=
\lambda_{\mathrm{len}}\,r_{\mathrm{len}}(\mathbf{a}_i)
+
\lambda_{\mathrm{range}}\,r_{\mathrm{range}}(\mathbf{a}_i)
\\
&\quad
+
\lambda_{\mathrm{schema}}\,r_{\mathrm{schema}}(\mathbf{a}_i),
\end{aligned}
\label{supp:eq_reward_format}
\end{equation}
where \(L_i = |\mathbf{a}_i|\), \(\mathcal{V}_{\mathrm{act}}\) denotes the valid
action-token vocabulary, and
\(\mathcal{V}_{\mathrm{act}} = \mathcal{V}_{\mathrm{scale}} \cup
\mathcal{V}_{\mathrm{traj}}\), where
\(\mathcal{V}_{\mathrm{scale}}\) and \(\mathcal{V}_{\mathrm{traj}}\) denote the
valid scale-token and trajectory-code-token vocabularies, respectively. We
define
% \begin{align}
% r_{\mathrm{len}}(\mathbf{a}_i)
% &=
% 1-\frac{|L_i-2S|}{2S},
% \label{supp:eq_reward_len}\\
% r_{\mathrm{range}}(\mathbf{a}_i)
% &=
% \frac{1}{\max(L_i,1)}
% \sum_{j=1}^{L_i}
% \mathbf{1}\!\left[a_{i,j}\in\mathcal{V}_{\mathrm{act}}\right],
% \label{supp:eq_reward_range}\\
% r_{\mathrm{schema}}(\mathbf{a}_i)
% &=
% \frac{1}{2S}
% \sum_{s=1}^{S}
% \Big(
% \mathbf{1}\!\left[a_{i,2s-1}\in\mathcal{V}_{\mathrm{scale}}\right]
% +
% \mathbf{1}\!\left[a_{i,2s}\in\mathcal{V}_{\mathrm{traj}}\right]
% \Big).
% \label{supp:eq_reward_schema}
% \end{align}
\begin{align}
r_{\mathrm{len}}(\mathbf{a}_i)
&=
1-\frac{|L_i-2S|}{2S},
\label{supp:eq_reward_len}\\
r_{\mathrm{range}}(\mathbf{a}_i)
&=
1-
\frac{1}{\max(L_i,1)}
\sum_{j=1}^{L_i}
\mathbf{1}\!\left[a_{i,j}\notin\mathcal{V}_{\mathrm{act}}\right],
\label{supp:eq_reward_range}\\
r_{\mathrm{schema}}(\mathbf{a}_i)
&=
1-
\frac{1}{2S}
\sum_{s=1}^{S}
\Bigg(
\mathbf{1}\!\left[
a_{i,2s-1}\notin\mathcal{V}_{\mathrm{scale}}
\right]
\notag\\
&\hspace{3.2em}
+
\mathbf{1}\!\left[
a_{i,2s}\notin\mathcal{V}_{\mathrm{traj}}
\right]
\Bigg),
\label{supp:eq_reward_schema}
\end{align}
where larger values of \(r_{\mathrm{len}}(\mathbf{a}_i)\),
\(r_{\mathrm{range}}(\mathbf{a}_i)\), and
\(r_{\mathrm{schema}}(\mathbf{a}_i)\) indicate fewer violations of the
action-token specification. Specifically,
\(r_{\mathrm{len}}(\mathbf{a}_i)\) decreases linearly with the normalized
deviation from the target action-tail length \(2S\);
\(r_{\mathrm{range}}(\mathbf{a}_i)\) is one minus the fraction of generated
tokens that fall outside the valid action-token vocabulary
\(\mathcal{V}_{\mathrm{act}}\); and
\(r_{\mathrm{schema}}(\mathbf{a}_i)\) is one minus the fraction of token
slots that violate the alternating scale-token and trajectory-code-token
layout required by the action-tokenization scheme. When generation
terminates early, missing slots are treated as schema violations.

In our implementation, we set
\(\lambda_{\mathrm{len}}=0.05\),
\(\lambda_{\mathrm{range}}=0.35\),
\(\lambda_{\mathrm{schema}}=0.60\), and
\(\alpha_{\mathrm{fmt}}=0.20\).
Because the action tail is fixed-length by design,
\(r_{\mathrm{len}}(\mathbf{a}_i)\) mainly captures deviations from the target
length and therefore acts as a light safeguard against malformed outputs, so
it is assigned a relatively small weight.
By contrast,
\(r_{\mathrm{range}}(\mathbf{a}_i)\) and
\(r_{\mathrm{schema}}(\mathbf{a}_i)\) directly penalize out-of-vocabulary
tokens and violations of the prescribed alternating token layout,
respectively.
Among them,
\(r_{\mathrm{schema}}(\mathbf{a}_i)\) receives the largest weight because
preserving the scale-token and trajectory-code-token ordering is most directly
related to whether the sampled sequence remains decodable under the frozen
action codebook.

Overall, the Stage-2 reward combines trajectory-space supervision for proposal
quality with token-space supervision for output validity. This design is
consistent with the action-token planning paradigm in the main text and
encourages the actor to generate candidate trajectories that are both safer to
execute and decodable under the frozen action codebook.

\section{Additional Experiments}
\label{sec:additional_Experiments}

\subsection{Implementation Details}
We adopt BEVFormer~\cite{li2024bevformer}, which features a pretrained ResNet-50 backbone and an FPN neck, as the BEV encoder. Spatial-temporal features are extracted via a single-layer Perception Transformer equipped with Temporal Self-Attention and MSDeformableAttention3D, sampling four points per pillar. In Stage-1, the model is optimized using the AdamW optimizer with an initial learning rate of $1\times 10^{-4}$, guided by a cosine learning-rate schedule and a warm-up ratio of 0.03. For Render-Teacher Alignment, the mask parameters are set to $\tau=0.2$ and $\gamma=2$. Within the SFT loss, the corresponding weights are defined as $\lambda_{\text{meta}}=1.0$, $\lambda_{\text{mask}}=0.02$, and $\lambda_{\text{attn}}=0.02$. In Stage-2, the optimization proceeds with a learning rate of $2\times 10^{-6}$. For each observation, a group of $K=3$ candidate trajectories is sampled. Furthermore, a clipping range of 0.18 is utilized, and a KL penalty coefficient of $\beta=0.02$ is applied. Finally, in Stage-3, the scoring head and the refinement head are trained via the AdamW optimizer with a learning rate of $3\times 10^{-4}$. The scoring head employs Smooth L1 regression alongside an additional ranking loss ($\lambda_{\text{rank}}=1.0$, $T_r=0.5$). The refinement head predicts a bounded residual with $\Delta_{\max}=(3\text{m}, 3\text{m}, 0.3\text{rad})$ and is optimized under Smooth L1 supervision.

\begin{figure}[t]
      \centering
       % 1. 压减图片与 Caption 之间的距离
      \setlength{\abovecaptionskip}{2pt} 
      % 2. 压减 Caption 与下方正文之间的距离
      \setlength{\belowcaptionskip}{-18pt}
      \includegraphics[width=\linewidth]{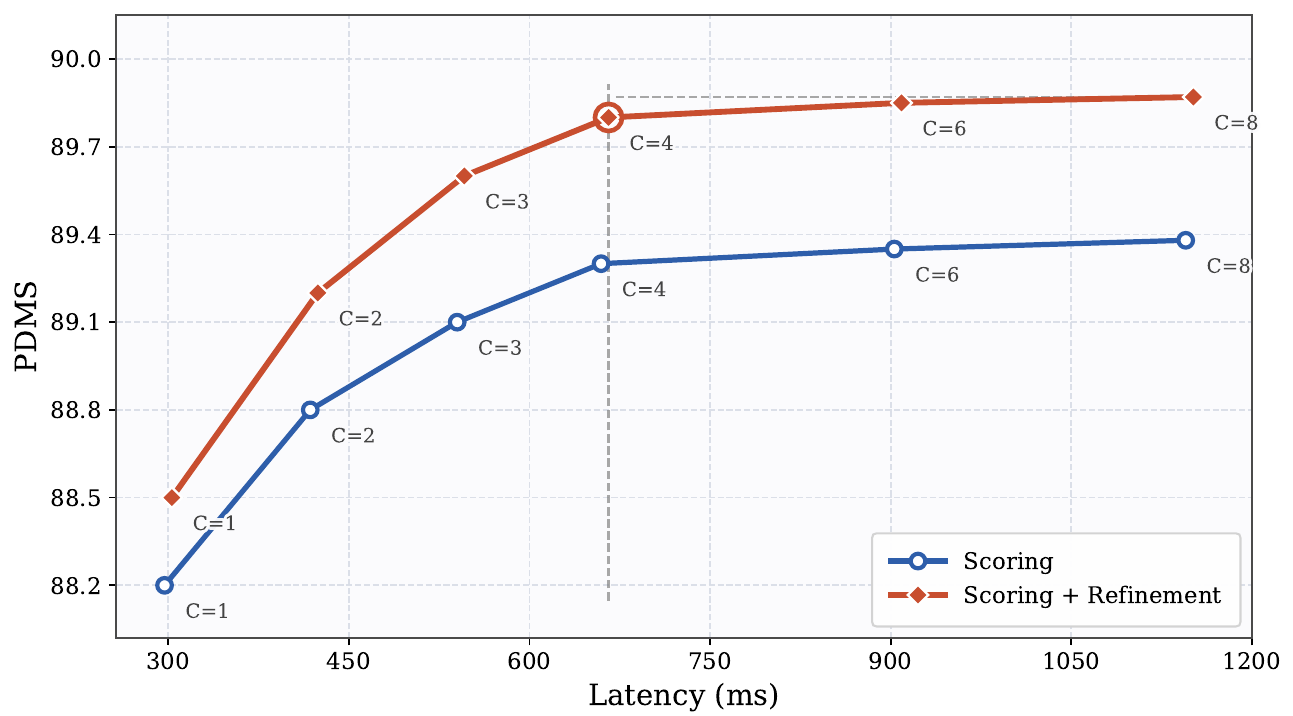}
      % \vspace{-1pt}
      \captionsetup{font={small}}
      \caption{Latency-PDMS Trade-off after SFT with Scoring and Refinement.
      }
      \label{fig:latency_tradeoff}
      % \vspace{-2pt} 
\end{figure}

\subsection{Ablation Studies}
\subsubsection{Latency Analysis.}
%解释一下耗时含义，第一次解码需要297ms耗时，其中bev encoder 45 ms cam encoder 20ms  第二次开始bev encoder和cam encoder不需要重复进行，且已经有了prefix prefill  测score和refine的速度  2ms和6ms  4条轨迹采样就能实现非常高的分数，并且提出一句AutoVLA~\cite{zhou2025autovla}和AutoDrive-P*3~\cite{ye2026autodrivetextp}分别耗时0.1Hz和0.5Hz，在能到达他们性能的前提下，the efficiency of the proposed framework surpasses that of previous end-to-end VLA models relying on an action codebook or text-based waypoints，并且做好对图示线的各种解释。
\cref{fig:latency_tradeoff} presents the trade-off between end-to-end latency and PDMS under different numbers of sampled trajectory candidates. The blue curve corresponds to trajectory sampling followed by the scoring head only, whereas the orange curve additionally applies the refinement head to the selected candidate. The latency of the first sampled candidate is 297 ms. This cost includes 45 ms for BEV encoding, 20 ms for camera encoding, and the initial decoder prefill over the full multimodal prefix before autoregressive action-token decoding. From the second candidate onward, the visual encoders are not executed again and the prefix prefill is already available. The critic introduces only a minor overhead, with about 2 ms for scoring and 6 ms for refinement. As a result, the gap between the two curves remains small, while the gain in PDMS is consistent across all candidate counts. More importantly, the performance saturates early: with four sampled candidates, the variant with scoring and refinement already reaches 89.8 PDMS, and further increasing the candidate count to six or eight yields only marginal gains. 
This behavior indicates that the proposed framework achieves a favorable operating point with a small candidate set, instead of relying on a large number of samples. At this practical operating point, the end-to-end latency remains below 700 ms, which is substantially more efficient than previous end-to-end VLA models that rely on an action codebook or text-based waypoints. In particular, AutoVLA~\cite{zhou2025autovla} and AutoDrive-$P^3$~\cite{ye2026autodrivetextp} operate at approximately 0.1 Hz and 0.5 Hz, respectively. Under a comparable performance regime, the efficiency of the proposed framework surpasses that of these prior methods.

\subsubsection{Effect of RFT Reward.}

Unless otherwise specified, all ablations in this subsection are conducted under
the Stage-2 reinforcement fine-tuning setting with the self-critic disabled. We
fix the GRPO group size to 3 in all Stage-2 ablation experiments. For the
driving-reward ablation, the full format reward is kept fixed. For the
format-reward ablation, the driving reward is fixed to the NC$+$TTC$+$DAC
configuration. Following the default Stage-2 setup, we use
\(\alpha_{\mathrm{fmt}}=0.20\) and
\((\lambda_{\mathrm{len}},\lambda_{\mathrm{range}},\lambda_{\mathrm{schema}})=(0.05,0.35,0.60)\)
for the format reward when it is enabled.

\begin{table}[h]
\caption{Ablation on the design of the Stage-2 driving reward.}
\label{tab:ablate_driving_reward}
\centering
\scriptsize
\setlength{\tabcolsep}{3.5pt}
\begin{tabular*}{\linewidth}{@{\extracolsep{\fill}}@{}c c c c c c c c >{\columncolor{gray!15}}c@{}}
\toprule
Exp. & NC & TTC & DAC & NC$\uparrow$ & DAC$\uparrow$ & TTC$\uparrow$ & EP$\uparrow$ & \textbf{PDMS}$\uparrow$ \\
\midrule
a & $\times$     & $\times$     & $\times$     & 98.7 & 97.0 & 94.7 & 84.5 & 89.5 \\
b & $\checkmark$ & $\times$     & $\times$     & \textbf{99.3} & 96.9 & 94.7 & 84.3 & 89.7 \\
c & $\checkmark$ & $\checkmark$ & $\times$     & 99.0 & 96.7 & 96.6 & 84.1 & 90.1 \\
d & $\checkmark$ & $\checkmark$ & $\checkmark$ & 99.2 & \textbf{97.6} & \textbf{96.9} & \textbf{84.2} & \textbf{90.5} \\
\bottomrule
\end{tabular*}
\end{table}

Table~\ref{tab:ablate_driving_reward} studies the effect of progressively
adding the three driving-reward terms under the Stage-2 setting. Starting from
the variant without any driving reward, adding NC alone substantially improves
NC and gives a modest gain in PDMS. Adding TTC on top of NC further improves
TTC and increases PDMS from 89.7 to 90.1, although NC and EP decrease
slightly. Adding DAC then gives the best overall result, with the highest PDMS
and the strongest balance across the planning sub-metrics. These results
suggest that NC, TTC, and DAC provide complementary supervision for collision
avoidance, safety margin, and drivable-area compliance, respectively. Their combination therefore provides the most balanced proposal-level training signal in Stage-2, and we adopt NC$+$TTC$+$DAC as the final driving-reward configuration.
Having identified the most effective driving-side supervision, we next examine
how the token-space format reward should be configured under this fixed driving
reward.

\begin{table}[tb]
\caption{Ablation on the design of the Stage-2 format reward.}
\label{tab:ablate_format_reward}
\centering
\scriptsize
\setlength{\tabcolsep}{3.0pt}
\begin{tabular*}{\linewidth}{@{\extracolsep{\fill}}@{}c c c c c c c c >{\columncolor{gray!15}}c@{}}
\toprule
Exp. & Len. & Range & Schema & NC$\uparrow$ & DAC$\uparrow$ & TTC$\uparrow$ & EP$\uparrow$ & \textbf{PDMS}$\uparrow$ \\
\midrule
a & $\times$     & $\times$     & $\times$     & 98.6 & 97.0 & \textbf{97.0} & 82.6 & 89.3 \\
b & $\checkmark$ & $\times$     & $\times$     & 98.7 & 97.1 & \textbf{97.0} & 83.0 & 89.5 \\
c & $\checkmark$ & $\checkmark$ & $\times$     & 98.9 & 97.3 & 96.9 & 83.6 & 89.9 \\
d & $\checkmark$ & $\checkmark$ & $\checkmark$ & \textbf{99.2} & \textbf{97.6} & 96.9 & \textbf{84.2} & \textbf{90.5} \\
\bottomrule
\end{tabular*}
\end{table}

Table~\ref{tab:ablate_format_reward} studies the design of the
Stage-2 format reward while keeping the driving reward fixed to NC$+$TTC$+$DAC.
Performance improves steadily as the format-reward components are added.
Using the Length term alone gives only a small gain over the baseline without
format reward, suggesting that it mainly acts as a mild auxiliary constraint.
Adding Range on top of Length brings a clearer improvement, and adding Schema
yields the best overall result in terms of PDMS. The gains are driven mainly
by EP, while NC and DAC improve more gradually and TTC changes little. This
suggests that the main role of the format reward is to maintain the validity
and structural consistency of the generated action-token sequences, which in
turn leads to more executable proposals and better planning quality. Overall,
the full Length$+$Range$+$Schema configuration provides the strongest format-side
supervision and is therefore used as the default Stage-2 setting.

\clearpage
\endgroup


%%%%%%%%%%%%%%%%%%%%%%%%%%%%%%%%%%%%%%%%%%%%%%%%%%%%%%%%%%%%%%%%%%%%%%%%%%%%%%%%

%%%%%%%%%%%%%%%%%%%%%%%%%%%%%%%%%%%%%%%%%%%%%%%%%%%%%%%%%%%%%%%%%%%%%%%%%%%%%%%%

%%%%%%%%%%%%%%%%%%%%%%%%%%%%%%%%%%%%%%%%%%%%%%%%%%%%%%%%%%%%%%%%%%%%%%%%%%%%%%%%

{
\scriptsize
\bibliographystyle{ieeeconf_2/IEEEtranBST/IEEEtran}
\bibliography{ref}
}

\end{document}